\documentclass[10pt,twocolumn,letterpaper]{article}

\usepackage{iccv}              

\usepackage{graphicx}
\usepackage{kotex}
\usepackage{enumitem}
\usepackage{multirow}
\usepackage{amsmath}
\usepackage{amssymb}
\usepackage{mathtools}
\usepackage{arydshln}
\usepackage{soul}
\usepackage{nicefrac}
\usepackage{booktabs}
\usepackage{stmaryrd}
\usepackage{setspace}
\usepackage{caption}
\usepackage{colortbl}
\usepackage{indentfirst}
\usepackage{booktabs}
\usepackage{multirow}
\usepackage{xcolor}
\usepackage{color}
\usepackage{array}
\usepackage{amsmath}
\usepackage{booktabs}
\usepackage{graphicx}
\usepackage{capt-of,etoolbox}
\usepackage{pifont}
\usepackage[accsupp]{axessibility}

\definecolor{iccvblue}{rgb}{0.21,0.49,0.74}
\usepackage[pagebackref,breaklinks,colorlinks,allcolors=iccvblue]{hyperref}

\newcommand\myfigure{
    \vspace{-1.7em}
    \centering
    \captionsetup{type=figure}
    \includegraphics[width=\textwidth]{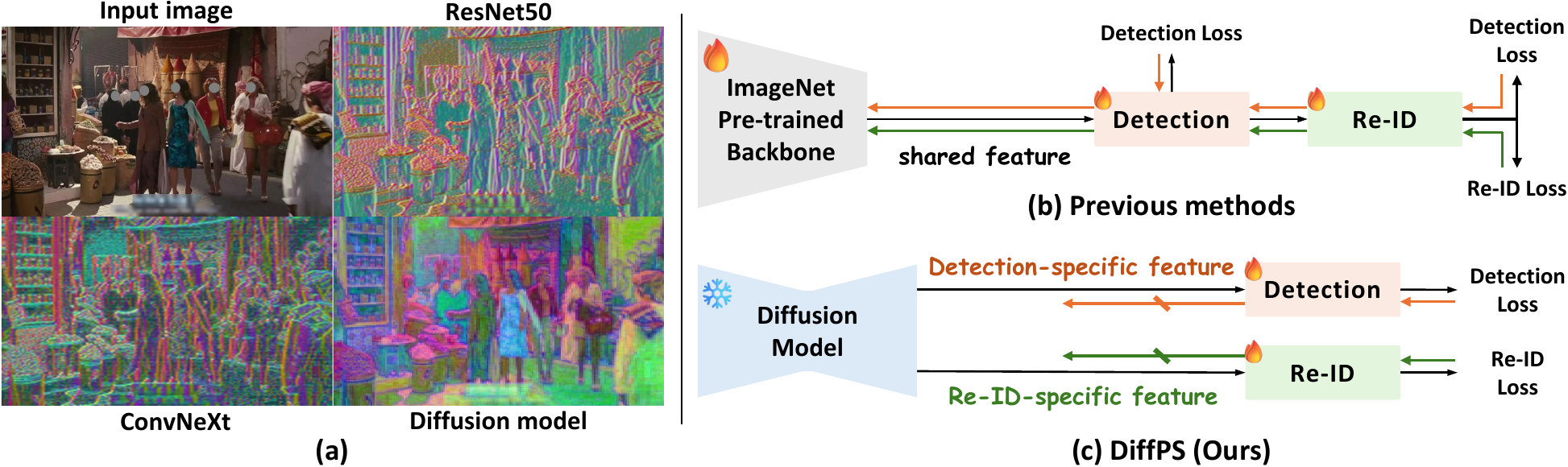}
    \vspace{-1.9em}
    \captionof{figure}{(a) PCA~\cite{PCA} visualization of feature maps, comparing widely-adopted ImageNet~\cite{ImageNet} pre-trained backbones (ResNet50~\cite{ResNet} and ConvNeXt~\cite{ConvNeXt}) in person search with a diffusion model. The diffusion features exhibit richer spatial context and fine-grained details. (b) Existing methods~\cite{OIM, DMRNet, NAE, SeqNet, AlignPS, PSTR, COAT, GFN, SEAS} share backbone features for detection and re-ID, causing optimization conflicts due to competing gradients (orange and green arrows). (c) Our DiffPS fully leverages diffusion priors, allowing us to freeze the backbone and utilize task-specific features for each sub-task, thereby preventing gradient interference and completely resolving the conflict.}
    \vspace{+1em}
    \label{fig:teaser}
    } 
\makeatletter
\apptocmd{\@maketitle}{{\myfigure{}\par}}{}{}
\makeatother

\def\paperID{9348} 
\def\confName{ICCV}
\def\confYear{2025}

\title{Leveraging Prior Knowledge of Diffusion Model for Person Search}

\author{
    Giyeol Kim$^1$\thanks{Equal contribution.}\quad
    Sooyoung Yang$^2$\footnotemark[1]\quad
    Jihyong Oh$^1$\quad
    Myungjoo Kang$^{2,3}$\quad
    Chanho Eom$^1$\thanks{Corresponding author.} \\[0.4em]
    $^1$GSAIM, Chung-Ang University \quad
    $^2$IPAI, Seoul National University \\
    $^3$Department of Mathematical Sciences and RIMS, Seoul National University \\[0.2em]
    {\small \texttt{\{giyeolkim, jihyongoh, cheom\}@cau.ac.kr, \{jimmy1016, mkang\}@snu.ac.kr}} \\[0.2em]
    \href{https://perceptualai-lab.github.io/DiffPS/}{\texttt{https://perceptualai-lab.github.io/DiffPS/}}
}

\begin{document}
\maketitle
\begin{abstract}
Person search aims to jointly perform person detection and re-identification by localizing and identifying a query person within a gallery of uncropped scene images. Existing methods predominantly utilize ImageNet pre-trained backbones, which may be suboptimal for capturing the complex spatial context and fine-grained identity cues necessary for person search. Moreover, they rely on a shared backbone feature for both person detection and re-identification, leading to suboptimal features due to conflicting optimization objectives. In this paper, we propose DiffPS (Diffusion Prior Knowledge for Person Search), a novel framework that leverages a pre-trained diffusion model while eliminating the optimization conflict between two sub-tasks. We analyze key properties of diffusion priors and propose three specialized modules: (i) Diffusion-Guided Region Proposal Network (DGRPN) for enhanced person localization, (ii) Multi-Scale Frequency Refinement Network (MSFRN) to mitigate shape bias, and (iii) Semantic-Adaptive Feature Aggregation Network (SFAN) to leverage text-aligned diffusion features. DiffPS sets a new state-of-the-art on CUHK-SYSU and PRW. 
\end{abstract}

\vspace{-1.5em}

\section{Introduction}

Person search aims to locate and identify a query person within a gallery of uncropped scene images, consisting of two main tasks: (i) \textit{person detection}~\cite{cao2021handcrafted, liu2019high, pang2019mask}, which localizes all person bounding boxes in each scene, and (ii) \textit{person re-identification (re-ID)}~\cite{liao2015person, zheng2017person, MGN, transreid}, which matches the detected person crops to the query. Early person search methods~\cite{MGTS, RDLR, IGPN}  tackle these tasks sequentially, using separate networks for detection and re-ID. While these methods achieve promising performance, their cascaded design leads to computational inefficiencies and prevents end-to-end inference. To address these limitations, recent methods~\cite{OIM, DMRNet, NAE, SeqNet, AlignPS, PSTR, COAT, GFN, SEAS} integrate person detection and re-ID within a unified framework, leveraging shared backbone features for both sub-tasks. This integration improves computational efficiency and enables end-to-end inference, making it the dominant paradigm in recent research.

Despite these advantages, recent methods still face two key challenges. The first challenge is the need for a backbone with strong generalization capabilities and rich prior knowledge to effectively support both person detection and re-ID. Existing methods~\cite{OIM, DMRNet, NAE, SeqNet, AlignPS, PSTR, COAT, GFN, SEAS} predominantly rely on ImageNet~\cite{ImageNet} pre-trained backbones (\textit{e.g.}, ResNet50~\cite{ResNet} or ConvNeXt~\cite{ConvNeXt}). However, as ImageNet mainly contains images with a single dominant object and simple backgrounds, these backbones are trained for category-level recognition, focusing on object presence~\cite{he2019rethinking, PretrainPS}. Therefore, they struggle to capture precise localization cues and instance-level discriminative features, both essential for person search in complex scenes with multiple overlapping individuals (Fig.~\ref{fig:teaser}~(a)). This limitation highlights the need for a pre-trained backbone capable of capturing rich contextual information and fine-grained details in challenging environments.

Another fundamental challenge is the optimization dilemma arising from the conflicting objectives between person detection and re-ID. In recent methods~\cite{OIM, DMRNet, NAE, SeqNet, AlignPS, PSTR, COAT, GFN, SEAS}, both sub-tasks share a common backbone, which is simultaneously optimized to serve two fundamentally different goals. Person detection aims to extract features that capture general human characteristics for distinguishing individuals from the background, while person re-ID requires highly discriminative features to differentiate between specific identities. This inherent conflict forces the shared backbone to learn contradictory representations, potentially degrading the performance of both sub-tasks~\cite{NAE, DMRNet, SeqNet, COAT, PSTR}. As shown in Fig.~\ref{fig:teaser}~(b), the backbone struggles to balance the two objectives, which hinders convergence and results in suboptimal performance due to incompatible updates within a single parameter space.

Recently, diffusion models~\cite{StableDiffusion, SD, GLIDE, DALL-E} have shown remarkable progress in image generation tasks. Trained on large-scale datasets (\textit{e.g.}, LAION-5B~\cite{LAION-5B}), they effectively capture both low-level visual features and high-level semantic relationships, enabling a comprehensive understanding of not only \textit{what} an object is but also \textit{where} it is located~\cite{diffusionseg} (Fig.\ref{fig:teaser}(a)). Leveraging these capabilities, pre-trained diffusion models have been successfully applied to a variety of fundamental vision tasks, \textit{e.g.}, image classification~\cite{li2023your, chen2024your}, segmentation~\cite{Diffumask, diffuse, nguyen2024dataset, li2023open, pnvr2023ld, xu2023open}, and semantic correspondence~\cite{Hyperfeature, Notall, zhang2024tale}. These successes highlight the potential of diffusion models as powerful general-purpose backbones, capable of handling diverse vision tasks even without fine-tuning. Inspired by these capabilities, the prior knowledge from a pre-trained diffusion model may appear well-suited for person search, due to its ability to capture both comprehensive spatial context and fine-grained details, which are essential for person search.

In this paper, we propose DiffPS (Diffusion Prior Knowledge for Person Search), a novel framework that fully exploits the prior knowledge of pre-trained diffusion models for person search (Fig.~\ref{fig:teaser}~(c)). To effectively harness the rich priors embedded in diffusion models, we first analyze their characteristics through four key properties: text condition, timesteps, hierarchical structure and shape bias. Based on these insights, we propose three specialized modules designed to maximize the diffusion priors for person search. First, we introduce a Diffusion-Guided Region Proposal Network (DGRPN), which refines person localization using cross-attention maps. Second, to mitigate the shape bias inherent in diffusion models, we propose a Multi-Scale Frequency Refinement Network (MSFRN) that enhances high-frequency details for improved identity discrimination. Finally, we design a Semantic-Adaptive Feature Aggregation Network (SFAN), which exploits the strong alignment between diffusion features and text embeddings to generate semantically enriched person representations. By fully leveraging diffusion priors, DiffPS eliminates the need for fine-tuning the backbone while maintaining strong performance. This allows us to freeze the backbone and extract task-specific features for person detection and re-ID, thereby completely resolving the inherent optimization conflict present in existing methods~\cite{OIM, NAE, SeqNet, COAT, PSTR, SEAS}. As a result, our framework enables independent optimization of both tasks, preventing mutual interference and ensuring that each branch is optimized for its respective objective. To validate the effectiveness of our approach, we conduct extensive qualitative and quantitative experiments, demonstrating that DiffPS not only outperforms existing methods but also establishes diffusion models as powerful backbones for person search. The main contributions of this work are:

\begin{itemize}
    \item We propose DiffPS, which effectively utilizes diffusion priors through three specialized modules, completely resolving the conflict between detection and re-ID.
    \item To the best of our knowledge, we are the first to leverage the prior knowledge of pre-trained diffusion models for person search.
    \item DiffPS achieves state-of-the-art performance on CUHK-SYSU~\cite{OIM} and PRW~\cite{PRW}.
\end{itemize}
\section{Related Work}

\subsection{Person Search}

Person search aims to localize and identify individuals within uncropped scene images. Early methods~\cite{MGTS, RDLR, IGPN} employ a two-stage approach using separate networks for detection and re-ID, while recent methods~\cite{OIM, DMRNet, NAE, SeqNet, AlignPS, PSTR, COAT, GFN, SEAS} integrate both tasks into a unified framework. However, these methods face limitations with ImageNet~\cite{ImageNet} pre-trained backbones, which struggle with contextual understanding and fine-grained feature extraction in complex scenes. To tackle this, SOLIDER~\cite{SOLIDER} introduces a pre-trained model that learns a general human representation through self-supervised learning on LUPerson~\cite{LUPerson} and PretrainPS~\cite{PretrainPS} adopts a hybrid learning paradigm across multiple datasets~\cite{CrowdHuman, Eurocity, MSMT17, CUHK03}. However, these models are primarily trained on cropped person images, which may limit their abilities to precisely locate individuals and adapt effectively across diverse and real-world scenes. In contrast, our approach leverages diffusion models pre-trained on diverse large-scale datasets, providing comprehensive spatial context and fine-grained details without domain constraints. 

Another fundamental challenge lies in the optimization conflict between detection and re-ID objectives. Previous methods~\cite{OIM, DMRNet, NAE, SeqNet, AlignPS, PSTR, COAT, GFN, SEAS} rely on a shared backbone for both sub-tasks, resulting in competing gradients that hinder optimization and degrade performance. In contrast, DiffPS leverages the rich prior knowledge of a pre-trained diffusion model, enabling us to freeze the backbone and utilize task-specific features, thereby avoiding the need for shared backbone optimization. Although DiffPS may appear conceptually similar to DMRNet~\cite{DMRNet} in its decoupled approach, it fundamentally differs by fully resolving the conflict through diffusion priors, enabling independent task optimization without any gradient interference.

\subsection{Diffusion Models for Downstream Tasks}
Recently, diffusion models~\cite{DDPM, GLIDE, SD, DALL-E, IMAGEN, StableDiffusion}, trained on large-scale datasets (\textit{e.g.}, LAION-5B~\cite{LAION-5B}), have demonstrated remarkable capabilities beyond image generation, extending to various vision understanding tasks. Several methods have leveraged diffusion features for segmentation~\cite{Diffumask, diffuse, li2023open, pnvr2023ld, xu2023open, diffusionseg, zhang2024tale} and classification~\cite{li2023your, chen2024your}. In addition, other methods~\cite{Notall, Hyperfeature} have focused on effective feature selection within diffusion models. These methods typically utilize diffusion features through simple concatenation or aggregation. In contrast, we maximize diffusion priors via our Multi-Scale Frequency Refinement Network, which enhances high-frequency details across multi-scale features. In person retrieval-related tasks, PSDiff~\cite{PSDiff} and DenoiseRep~\cite{DenoiseRep} incorporate diffusion and denoising processes, but significantly differ from our approach. Specifically, PSDiff~\cite{PSDiff} adapts denoising algorithms for bounding box regression similar to DiffusionDet~\cite{diffusiondet}, and DenoiseRep~\cite{DenoiseRep} applies denoising techniques in representation learning. These methods do not leverage the internal representations of pre-trained diffusion models trained on large-scale datasets. In contrast, our method fully exploits the prior knowledge embedded in a pre-trained diffusion model, effectively adapting it for person search.
\section{Method}
\label{sec:label}
In this section, we first introduce the diffusion model and its UNet~\cite{UNet} architecture (Sec.~\ref{sec:Preliminaries}), followed by an analysis of diffusion priors for person search (Sec.~\ref{sec:Prior Knowledge of Diffusion Model}). Finally, we present our proposed framework (Sec.~\ref{sec:Framework}).

\subsection{Preliminaries}
\label{sec:Preliminaries}

\paragraph{Diffusion models.}
Diffusion models~\cite{DDPM, DDIM, SD, StableDiffusion} generate images by iteratively refining Gaussian noise through a forward and reverse process. In the forward process, noise is gradually added to a clean image $\mathbf{x}_0$ over $T$ timesteps following a noise schedule $\{\alpha_t\}_{t=1}^{T}$, transforming it into a nearly pure noise sample $\mathbf{x}_T$ as $\mathbf{x}_t = \sqrt{\Bar{\alpha}_t} \mathbf{x}_0 + \sqrt{1 - \Bar{\alpha}_t} \epsilon$, where $\epsilon \sim \mathcal{N}(0, \mathbf{I})$ and $\Bar{\alpha}_t = \prod_{k=1}^{t} \alpha_k$. The reverse process removes noise using a denoising network $\mathcal{F}_\theta$, estimating $\epsilon \approx \mathcal{F}_\theta(\mathbf{x}_t, t)$ to reconstruct the image. For text-conditioned generation, CLIP~\cite{CLIP} text encoder $\mathcal{T}(\cdot)$ converts a prompt $\mathbf{p}$ into an embedding $\mathbf{T}_p = \mathcal{T}(\mathbf{p}) \in \mathbb{R}^{77\times d}$, guiding the denoising process via cross-attention. 

\vspace{-0.9em}

\begin{figure}[t!]
  \centering
  \includegraphics[width=\linewidth]{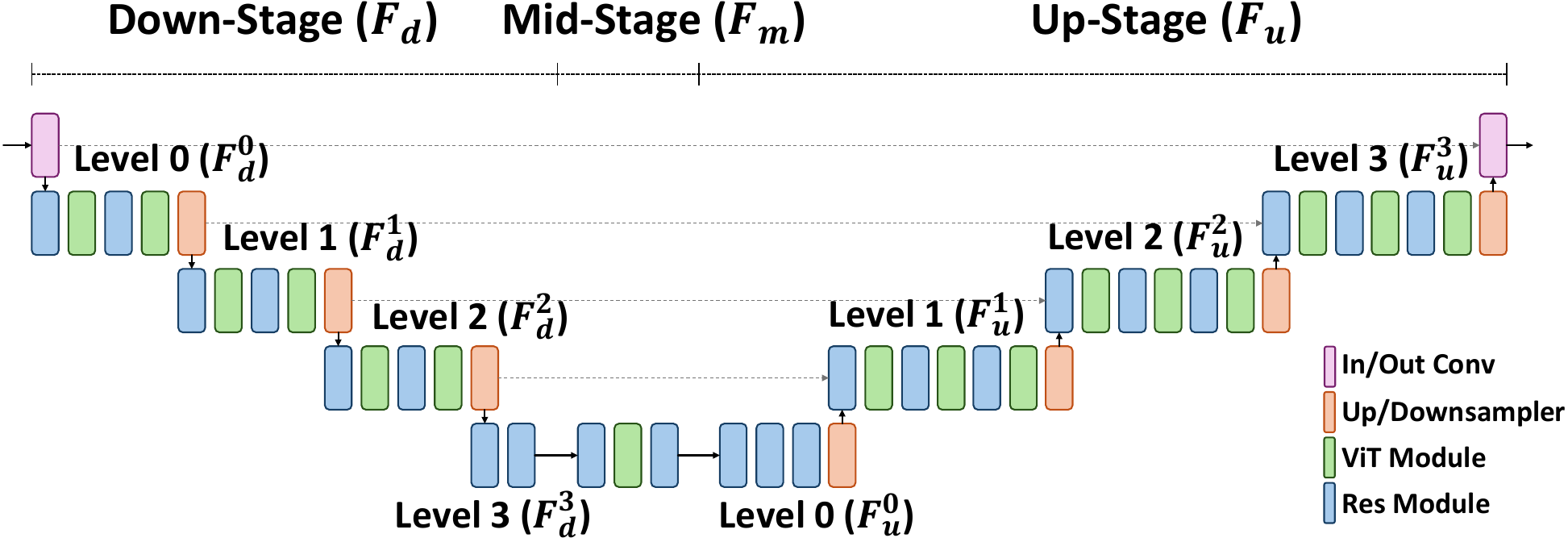}
  \vspace{-1.5em}
  \caption{Detailed architecture of the UNet~\cite{UNet} in diffusion models~\cite{SD, StableDiffusion}, comprising down-stage, mid-stage, and up-stage for hierarchical feature processing. (Best viewed in color.)}
  \label{fig:UNet}
  \vspace{-1em}
\end{figure}

\vspace{-0.3em}
\paragraph{UNet Architecture in Diffusion Models.}  
The UNet~\cite{UNet} architecture in diffusion models~\cite{SD, StableDiffusion} follows a hierarchical structure comprising three main stages, as shown in Fig.~\ref{fig:UNet}. The down-stage progressively reduces spatial resolution while increasing channel dimensions, whereas the up-stage restores resolution using skip connections from corresponding down-stage levels. Both the down-stage and up-stage consist of four resolution levels, each containing multiple sequential processing blocks for feature refinement. Between these stages, the mid-stage operates on the most compressed features at the bottleneck. We denote the features extracted at level $l$ of the down-stage as $\mathbf{F}_{d}^{l}$ and those from the corresponding up-stage level as $\mathbf{F}_{u}^{l}$. Further details are provided in the Suppl.~\textcolor{iccvblue}{A}.

\subsection{Diffusion Priors for Person Search}
\label{sec:Prior Knowledge of Diffusion Model}

\begin{figure}[t!]
   \centering
   \includegraphics[width=0.90\linewidth, height=10.2cm]{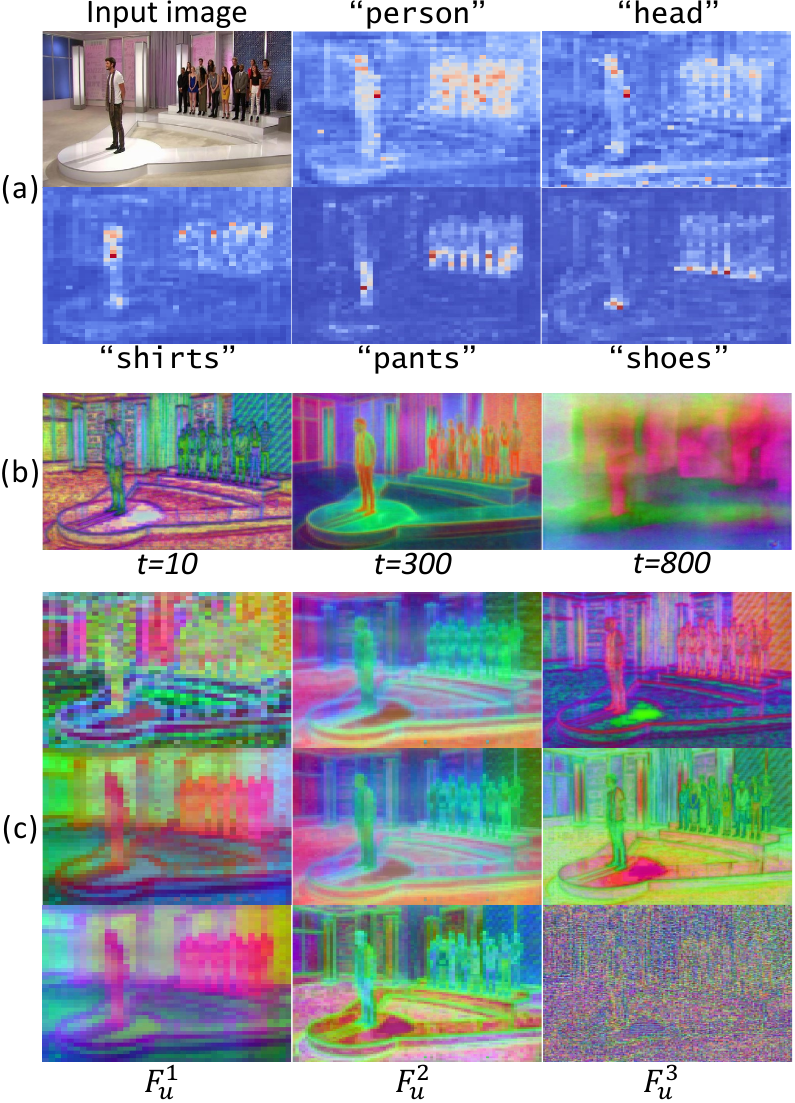}
   \vspace{-0.7em}
   \caption{(a) Cross-attention maps highlighting different semantic regions based on textual queries. (b) PCA~\cite{PCA} visualization of features extracted from the ViT~\cite{ViT} block of \(\mathbf{F}_u^3\) at different timesteps (\textit{t}). (c) PCA visualization of up-stage feature maps (\( \mathbf{F}_u^1, \mathbf{F}_u^2, \mathbf{F}_u^3 \)), with columns representing different up-stage levels and rows corresponding to features from ViT blocks within each level.} 
   \label{fig:prior}
   \vspace{-1.5em}
\end{figure}

\paragraph{Text condition.}
Pre-trained diffusion models~\cite{SD, StableDiffusion, IMAGEN} exhibit a strong alignment between image features and text embeddings via cross-attention~\cite{xu2023open, VPD, pnvr2023ld, Genpromp}. As shown in Fig.~\ref{fig:prior} (a), attention maps for different textual tokens highlight relevant image regions. This alignment provides a strong semantic prior for person localization and enables the extraction of discriminative features for body parts and clothing. By leveraging these features, the model effectively suppresses background noise and mitigates occlusions, ensuring a more robust representation for person search.
\vspace{-0.8em}

\vspace{-0.6em}
\paragraph{Timesteps.}
The effectiveness of diffusion features varies significantly depending on the timestep at which they are obtained, as shown in Fig.~\ref{fig:prior}~(b). In datasets~\cite{ImageNet, coco} where objects are sparse and easily distinguishable, features extracted at early-to-middle timesteps (\textit{e.g.,} \textit{t}=300) tend to be the most informative. At very early timesteps (\textit{e.g.,} \textit{t}=10), the model removes only minimal noise, making feature extraction too trivial to capture meaningful representations. Conversely, at later timesteps (\textit{e.g.,} \textit{t}=800), excessive noise forces the model to focus more on denoising rather than preserving fine-grained details, leading to a degradation in feature quality~\cite{mukhopadhyay2024text, VPD, Genpromp}. However, this pattern may differ in the person search, which involves cluttered backgrounds, multiple objects, and diverse real-world conditions. These images exhibit high visual complexity even before artificial noise is introduced, making it harder to extract clear and discriminative features. Additionally, real-world noise from camera artifacts, motion blur, and lighting variations further intensifies the challenge. In such cases, adding synthetic noise exacerbates the difficulty of distinguishing meaningful features from irrelevant information. As a result, extracting features at earlier timesteps may be more beneficial (Fig.~\ref{fig:prior}~(b)), as they preserve finer detailes while minimizing interference from both synthetic and real-world noise. This tendency further supported by our empirical results.

\vspace{-0.9em}
\paragraph{Hierarchical structure.}
The UNet~\cite{UNet} architecture in diffusion models follows a hierarchical structure (Fig.~\ref{fig:UNet}). In particular, the up-stage restores spatial information by integrating local features from the down-stage via skip connections while incorporating global context from the mid-stage. Consequently, up-stage features tend to be more informative than those from earlier stages, as they retain both fine-grained details and high-level semantics~\cite{VPD, Notall, Hyperfeature}, as shown in Fig.~\ref{fig:prior}~(c). In person search, as detection requires spatial precision and re-ID relies on distinctive features, the up-stage features may appear well-suited for both tasks. However, not all layers within the up-stage may be equally beneficial. Since the UNet is originally designed for noise prediction, its final layers (Fig.~\ref{fig:prior}~(c), row 3, column 3) tend to focus on estimating noise maps rather than extracting meaningful features. Furthermore, even within the same hierarchical level, different layers exhibit distinct feature properties due to variations in convolutional operations, attention mechanisms, and residual connections. Our experiments confirm that a careful selection of specific up-stage layers significantly improves person search performance.

\vspace{-0.9em}
\paragraph{Shape bias.}
Pre-trained diffusion models~\cite{SD, StableDiffusion, IMAGEN, DALL-E} have demonstrated superior discriminative power across various downstream tasks~\cite{Hyperfeature, zhang2024tale, pnvr2023ld, li2023open, xu2023open, chen2024your} compared to traditional discriminative models~\cite{ResNet, ConvNeXt, ViT}. This advantage stems from their ability to capture both global context and fine-grained features. Notably, diffusion features naturally emphasize global structures due to their progressive denoising process, which reconstructs images from noise by first recovering low-frequency components before refining finer details. While this property enhances robustness and improves high-level feature alignment, it may also introduce a shape bias, where low-frequency components tend to be more dominant than fine textures~\cite{intriguing, Diffusion_sketch}. For person search, where both global structure and fine-grained details are essential, leveraging diffusion features while further enhancing high-frequency information can lead to even greater discriminative power~\cite{PHA, wang2022domain}.

\begin{figure*}[t!]
    \centering
    \includegraphics[width=\textwidth]{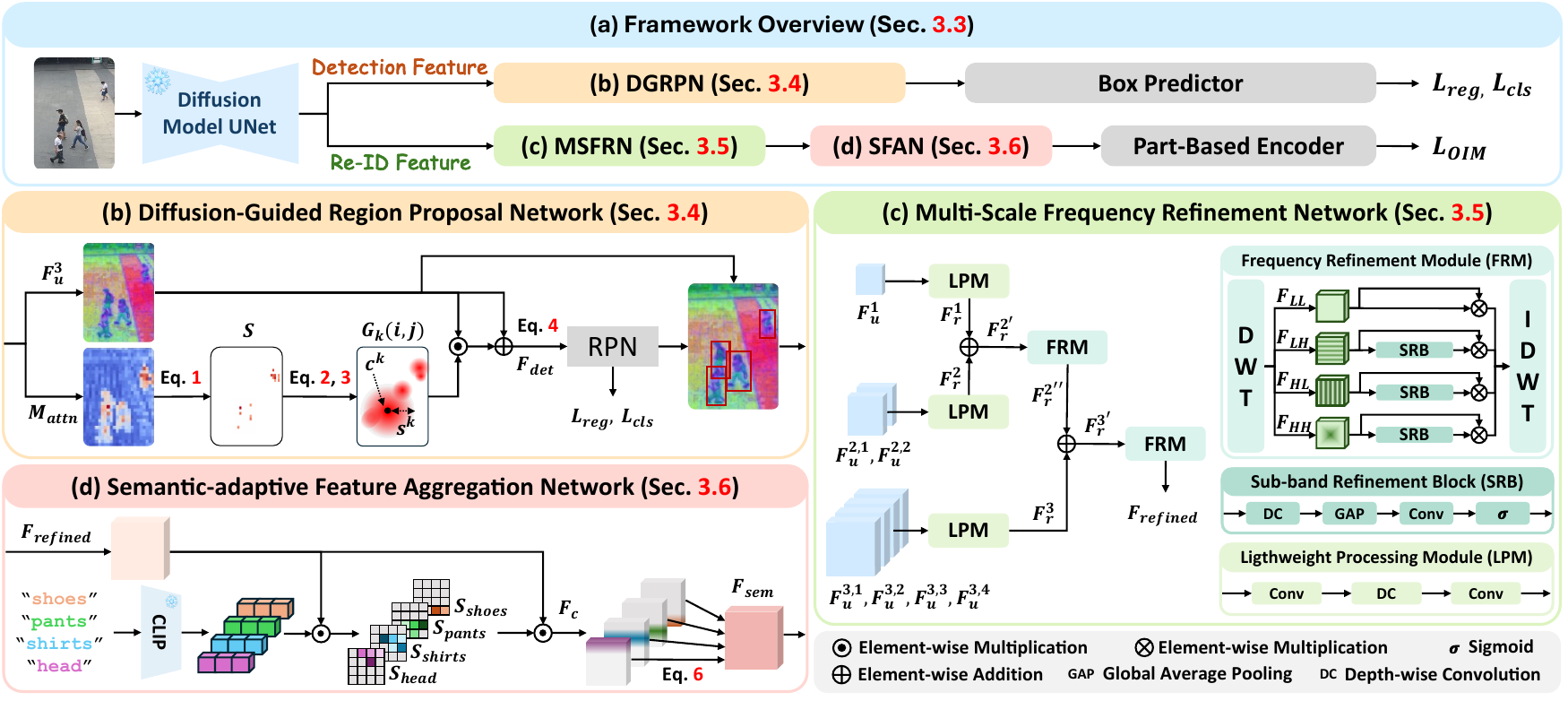}
    \vspace{-1.85em}
    \caption{(a) Overview of DiffPS framework. DiffPS leverages a pre-trained diffusion model's UNet as the backbone, with three specialized modules: (b) DGRPN refines region proposals using cross-attention maps, (c) MSFRN enhances high-frequency details via multi-scale frequency refinement, and (d) SFAN incorporates text-aligned semantic features for re-ID. (Best viewed in color.)}
    \label{fig:Framework}
    \vspace{-1.2em}
\end{figure*}

\subsection{Framework Overview}
\label{sec:Framework}
The overall architecture of our DiffPS is illustrated in Fig.~\ref{fig:Framework}~(a). DiffPS comprises three main components: a frozen pre-trained diffusion model backbone and two branches for detection and re-ID. Through empirical analysis (Table~\ref{tab:feature_selection}), we found that the detection branch can be effectively achieved using only a single feature map and a cross-attention map, while the re-ID branch benefits from leveraging multiple feature maps to capture discriminative details. Both branches are carefully designed to maximize the prior knowledge embedded in the pre-trained diffusion model. For the detection branch, we adopt Faster R-CNN~\cite{Faster-RCNN}, following previous methods~\cite{OIM, OIM++, SEAS, COAT, GFN, SeqNet, QEEPS, DMRNet}. However, instead of a conventional Region Proposal Network (RPN)~\cite{Faster-RCNN}, we introduce a Diffusion-Guided Region Proposal Network (DGRPN), which utilizes diffusion priors to guide the localization of potential person regions. In the re-ID branch, we first refine the diffusion features using a Multi-Scale Frequency Refinement Network (MSFRN) to enhance high-frequency details which are critical for identity discrimination. The refined features are then processed via RoI-Align~\cite{Faster-RCNN}, and we exploit the strong alignment between diffusion features and text embeddings by incorporating Semantic-adaptive feature aggregation network (SFAN). We further adopt a simple stripe-based partitioning network, which is widely used in re-ID methods~\cite{SEAS, MGN, su2017pose, ISGAN, sun2018beyond}, to extract the final person representation. Our decoupled structure allows for flexible architecture choices, enabling any detection or re-ID module to be seamlessly integrated in a plug-and-play manner, as demonstrated in the Suppl.~\textcolor{iccvblue}{B}. For training, we employ standard loss functions: Smooth-L1 loss and cross-entropy loss for detection, and Online Instance Matching (OIM)~\cite{OIM} loss for re-ID. In the following sections, we provide detailed explanations of each module in our framework.

\subsection{Diffusion-Guided Region Proposal Network} 
DGRPN leverages the strong alignment between diffusion features and text embeddings by utilizing the cross-attention mechanism associated with the \texttt{"person"} token embedding, as discussed in Sec.~\ref{sec:Prior Knowledge of Diffusion Model}. Specifically, we extract the feature map $\mathbf{F}_{u}^{3}$ and the cross-attention map $\mathbf{M}_{att}$ from the corresponding cross-attention layer. The cross-attention map encodes attention scores that highlight person-related regions. However, due to the complexity of person search datasets with their cluttered backgrounds and multiple overlapping individuals, this attention map may lack sharpness or be imprecise. To address this, we first threshold the cross-attention map to retain only high-confidence regions:
\vspace{-0.2em}
\begin{equation}    
    \mathbf{M}_{th}(i,j) = 
    \begin{cases}
      \mathbf{M}_{att}(i,j), & \text{if } \mathbf{M}_{att}(i,j) > \tau \\
      0 & \text{otherwise},
    \end{cases}
    \vspace{-0.2em}
\end{equation}
where $\tau$ is a predefined threshold. Next, we use the set of sampled pixels $\mathcal{S} = \{(i, j) \mid \mathbf{M}_{th}(i,j) > 0\}$ as Gaussian centers. Each selected pixel $(i,j) \in \mathcal{S}$ serves as a candidate center $(c_x^{k}, c_y^{k})$ for a Gaussian distribution, where $k$ indexes different detected peaks. For each sampled pixel, we define its Gaussian standard deviation $(s_w^{k}, s_h^{k})$ based on local spatial statistics:
\vspace{-0.3em}
\begin{equation}
    s_w^{k} = \max\left(\delta, \sqrt{\sum_{i,j \in \mathcal{N}(c_x^{k}, c_y^{k})} (i - c_x^{k})^2 \cdot \mathbf{M}_{th}(i,j)}\right),
\end{equation}
where $\mathcal{N}(c_x^{k}, c_y^{k})$ represents a local neighborhood around $(c_x^{k}, c_y^{k})$, and $\delta$ is a hyperparameter ensuring a minimum size for the Gaussian. The standard deviation $s_h^{k}$ is computed in the same manner for the vertical axis. Using these centers and variances, we construct Gaussian maps:
\begin{equation}
    G_k(i,j) = \exp\left(-\frac{(i - c_x^{k})^2}{\beta (s_w^{k})^2} - \frac{(j - c_y^{k})^2}{\beta (s_h^{k})^2}\right),
\end{equation}
where $\beta$ is a learnable scaling factor. Since multiple peaks can exist in the attention map, we aggregate all Gaussian maps by taking the element-wise maximum across them, producing the final map \( G_{\text{det}}(i,j) = \max_k G_k(i,j) \).
Finally, we modulate the detection-specific feature map $\mathbf{F}_{u}^{3}$ using the generated Gaussian map as follows:
\vspace{-0.3em}
\begin{equation}
    \mathbf{F}_{\text{det}} = \mathbf{F}_{u}^{3} + \gamma \left( G_{\text{det}} \odot \mathbf{F}_{u}^{3} \right),
    \vspace{-0.3em}
\end{equation}
where $\odot$ denotes element-wise multiplication, and $\gamma$ is a learnable parameter that controls the influence of the Gaussian-modulated features. This enhanced feature map $\mathbf{F}_{\text{det}}$ is used as input to the region proposal network.

\subsection{Multi-Scale Frequency Refinement Network} 
As discussed in Sec.~\ref{sec:Prior Knowledge of Diffusion Model}, diffusion features effectively capture both global context and fine-grained details. However, they tend to exhibit a shape bias due to the denoising process. To mitigate this, we propose MSFRN to improve discriminability by enhancing high-frequency details in diffusion features. Specifically, we first extract re-ID specific multi-scale feature maps from the up-stage, using four feature maps from $\mathbf{F}_{u}^{3}$, two from $\mathbf{F}_{u}^{2}$, and one from $\mathbf{F}_{u}^{1}$, yielding a total of seven feature maps. Then, MSFRN processes each feature independently to capture unique characteristics. To be specific, for each feature map $\mathbf{F}_{u}^{l, i}$, where $i$ indexes different layers within the same level $l$, we apply a Lightweight Processing Module (LPM) consisting of $1 \times 1$ and depth-wise convolutions: $\mathbf{F}_{r}^{l, i} = \text{Conv}_{1 \times 1} \left(\text{DepthConv} \left(\text{Conv}_{1 \times 1}(\mathbf{F}_{u}^{l, i})\right)\right).$ These processed features are then aggregated within each level using concatenation followed by a $1 \times 1$ convolution, yielding three multi-scale refined features $\mathbf{F}_{r}^{3}, \mathbf{F}_{r}^{2},$ and $\mathbf{F}_{r}^{1}$.

To enhance high-frequency details, we adopt a hierarchical frequency decomposition strategy within the Frequency Refinement Module (FRM). The lowest-resolution feature $\mathbf{F}_{r}^{1}$ is first upsampled and added to $\mathbf{F}_{r}^{2}$, followed by Layer Normalization to yield $\mathbf{F}_{r}^{2'}$. We then apply the Discrete Wavelet Transform (DWT) to $\mathbf{F}_{r}^{2'}$, obtaining four sub-bands: a low-frequency component $\mathbf{F}_{LL}$ and three high-frequency components $\mathbf{F}_{LH}, \mathbf{F}_{HL}, \mathbf{F}_{HH}$. Each high-frequency band is refined using a Sub-band Refinement Block (SRB), consisting of depth-wise separable convolution, global average pooling, a $1 \times 1$ convolution, and a sigmoid activation to produce a channel-wise attention vector $\mathbf{s}_X$. This vector modulates the feature as: $\hat{\mathbf{F}}_{X} = \mathbf{s}_X \cdot \text{DepthConv}(\mathbf{F}_{X}), \quad X \in \{LH, HL, HH\}.$ To reconstruct the refined feature, we apply the Inverse DWT with learnable scaling factors $\gamma_X$:
\begin{equation}
    \mathbf{F}_{r}^{2''} = \text{IDWT}(\mathbf{F}_{LL}, \gamma_{LH} \hat{\mathbf{F}}_{LH}, \gamma_{HL} \hat{\mathbf{F}}_{HL}, \gamma_{HH} \hat{\mathbf{F}}_{HH}).
\end{equation}
This output is upsampled and added to $\mathbf{F}_{r}^{3}$, where the same process is repeated. Finally, a $1 \times 1$ convolution is applied to the refined feature $\mathbf{F}_{r}^{3''}$ to obtain the final output $\mathbf{F}_{\text{refined}}$.

Our MSFRN effectively enhances fine-grained discriminative information by leveraging hierarchical frequency decomposition and multi-scale fusion, mitigating the loss of high-frequency details in shape-biased features.

\subsection{Semantic-Adaptive Feature Aggregation Network}
SFAN harnesses the strong alignment between diffusion features and text embeddings in their shared semantic space to enhance person representations. Given the refined feature map $\mathbf{F}_{\text{refined}}$ obtained from MSFRN, we first compute the similarity between each spatial feature and pre-defined text embeddings corresponding to human body regions, including \texttt{"head"}, \texttt{"shirts"}, \texttt{"pants"}, and \texttt{"shoes"}. Through experiments, we identify the optimal combination of text prompts, with additional details provided in the Suppl.~\textcolor{iccvblue}{C}. These text embeddings, extracted from the CLIP~\cite{CLIP} text encoder of a text-to-image diffusion model~\cite{SD, StableDiffusion}, are computed only once. The cosine similarity between the refined feature map and each text embedding token is computed, producing four spatial semantic maps, $\mathbf{S}_{\text{head}}, \mathbf{S}_{\text{shirts}}, \mathbf{S}_{\text{pants}}, \mathbf{S}_{\text{shoes}}$, where each $\mathbf{S}_{c}$ highlights the relevance of each feature location to the corresponding body part $c$. We utilize these semantic maps to generate spatially adaptive person feature maps. Specifically, each semantic map $\mathbf{S}_{c}$ is normalized via a softmax function over all categories to obtain a probability distribution $\hat{\mathbf{S}}_{c}$. The refined feature map is then weighted by these normalized maps to emphasize semantic regions while preserving global spatial structure. This process is formulated as $\mathbf{F}_{c} = \hat{\mathbf{S}}_{c} \odot \mathbf{F}_{\text{refined}}$, where $\odot$ denotes element-wise multiplication, applying the semantic relevance as a spatial attention mechanism. To integrate all semantic regions while maintaining spatial consistency, the refined body-part features are aggregated as:
\vspace{-0.4em}
\begin{equation}
    \mathbf{F}_{\text{sem}} = \sum_{c} \mathbf{W}_{c} \mathbf{F}_{c},
    \vspace{-0.6em}
\end{equation}
where $\mathbf{W}_{c}$ is a learnable weight parameter that adaptively balances contributions from different body regions. Through this adaptive weighting mechanism, SFAN emphasizes informative body regions while suppressing irrelevant background noise and occluded regions.

\section{Experiments}
\label{sec:experiments}

\subsection{Experimental details}

\paragraph{Datasets and evaluation metrics.} \label{sec:Implementation details} Following prior methods~\cite{OIM, DMRNet, NAE, SeqNet, AlignPS, PSTR, COAT}, we evaluate our model on CUHK-SYSU~\cite{OIM} and PRW~\cite{PRW}. CUHK-SYSU contains 18,184 images, 96,143 pedestrian boxes, and 8,432 identities, with 11,206 images for training and 6,978 for testing. PRW consists of 11,816 frames from six cameras, with 43,110 boxes for 932 identities, including 5,704 training and 6,112 test images. Performance is measured using mean average precision (mAP) and top-1 accuracy (Top-1).
\vspace{-1em}

\paragraph{Implementation details.} We train our model end-to-end for 20 epochs. We use the Adam~\cite{Adam} optimizer, where $\beta_1$ and $\beta_2$ are set to 0.9 and 0.999, respectively. We use a warm-up and step decay strategy, linearly increasing the learning rate from a starting point of $1 \times 10^{-7}$ to $1 \times 10^{-4}$ over the first epoch. The batch size is set to 5 and random horizontal flip is used as the augmentation method. In DGRPN, the threshold \(\tau\) is set to 0.7 and $\delta$ to 5. We use Stable Diffusion~\cite{StableDiffusion} v2-1 as the backbone. For the part-based encoder, we follow the design of SEAS~\cite{SEAS}.

\vspace{-0.5em}

\begin{table}[t!]
\centering
\resizebox{\linewidth}{!}{
\footnotesize
\begin{tabular}{l|l|cc|cc}
\toprule
\multirow{2}{*}{Method} & \multirow{2}{*}{Backbone} & \multicolumn{2}{c|}{CUHK-SYSU} & \multicolumn{2}{c}{PRW} \\
\cmidrule(lr){3-4} \cmidrule(lr){5-6}
       &       & mAP   & Top-1 & mAP & Top-1  \\
\midrule \midrule

OIM~\cite{OIM}      & ResNet50    & 75.5 & 78.7  & 21.3 & 49.4 \\
IAN~\cite{IAN}      & ResNet50    & 76.3 & 80.1  & 23.0 & 61.9 \\
QEEPS~\cite{QEEPS}     & ResNet50    & 88.9 & 89.1  & 37.1 & 76.7 \\
BINet~\cite{BINet}      & ResNet50    & 90.0 & 90.7  & 45.3 & 81.7 \\
APNet~\cite{APNet}      & ResNet50    & 88.9 & 89.3  & 41.9 & 81.4 \\
NAE~\cite{NAE}      & ResNet50    & 91.5 & 92.4  & 43.3 & 80.9 \\
NAE+~\cite{NAE}     & ResNet50    & 92.1 & 92.9  & 44.0 & 81.1 \\
PGSFL~\cite{PGSFL}   & ResNet50    & 90.2 & 91.8  & 42.5 & 83.5 \\
SeqNet~\cite{SeqNet}      & ResNet50    & 93.8 & 94.6  & 46.7 & 83.4 \\
DMRNet~\cite{DMRNet}   & ResNet50    & 93.2 & 94.2  & 46.9 & 83.3 \\
AlignPS~\cite{AlignPS}      & ResNet50    & 93.1 & 93.4  & 45.9 & 81.9 \\
COAT~\cite{COAT}   & ResNet50 &94.2 &94.7 &53.3 &87.4 \\
PSTR~\cite{PSTR}   & ResNet50   & 93.5 &  95.0  & 49.5 &  87.8 \\
PSTR~\cite{PSTR}   & PVTv2-B2    & 95.2 & 96.2  & 56.5 & 89.7 \\
SeqNeXt~\cite{GFN} & ConvNeXt &96.1 &96.5 &57.6 &89.5 \\
SeqNeXt+GFN~\cite{GFN} & ResNet50 &94.7 &95.3 &51.3 &90.6 \\
SeqNeXt+GFN~\cite{GFN} & ConvNeXt &96.4 &97.0 &58.3 &\textbf{92.4} \\
SOLIDER~\cite{SOLIDER} & Swin-S &95.5 &95.8 &59.8 &86.7 \\
SEAS~\cite{SEAS}    & ResNet50 &96.2 &97.1 &52.0 &85.7 \\
SEAS~\cite{SEAS}    & ConvNeXt & \underline{97.1} &\underline{97.8} & \underline{60.5} &89.5 \\
\midrule
DiffPS (Ours) & SD v2-1  & \textbf{97.8} &  \textbf{98.4}  & \textbf{62.0} &  \underline{91.0} \\
\bottomrule
\end{tabular}
}
\vspace{-0.7em}
\caption{Comparison with the state-of-the-art methods on CUHK-SYSU~\cite{OIM} and PRW~\cite{PRW} test sets. Numbers in bold indicate the best performance and underscored ones are the second best.}
\label{tab:comparison to state-of-the-art}
\vspace{-1.2em}
\end{table}

\subsection{Comparison to the state-of-the-arts}
\label{sec:Comparison to the state-of-the-arts}
\paragraph{CUHK-SYSU.} On the CUHK-SYSU~\cite{OIM} dataset, our DiffPS achieves state-of-the-art performance with 97.8\% mAP and 98.4\% Top-1 accuracy, surpassing all previous methods. Notably, DiffPS outperforms ResNet50~\cite{ResNet}-based methods by a large margin  of 1.6-22.3\% in mAP. Even when compared to methods using more advanced backbones like ConvNeXt~\cite{ConvNeXt} and Swin-S~\cite{swintransformer}, DiffPS demonstrates superior performance, highlighting the effectiveness of leveraging the diffusion priors for person search.

\vspace{-1.2em}
\paragraph{PRW.} The PRW dataset presents a more challenging scenario, yet DiffPS maintains its superior performance with 62.0\% mAP and 91.0\% Top-1 accuracy. This represents a significant improvement of 2.2\% mAP over SOLIDER~\cite{SOLIDER} and 1.5\% over SEAS~\cite{SEAS} with ConvNeXt~\cite{ConvNeXt}. While our Top-1 accuracy is slightly lower than SeqNeXt+GFN~\cite{GFN}, we achieve a significantly higher mAP.

\subsection{Ablation Study}
\label{sec:Ablation Study}

\paragraph{Resolving conflict.}
As shown in Table~\ref{tab:joint optimization}, we compare our method with DMRNet~\cite{DMRNet} and DMRNet++~\cite{DMRNet++}, which also employ a decoupled design but still suffers from optimization conflicts. These methods exhibits a noticeable performance degradation when jointly optimizing detection and re-ID tasks compared to their individual optimization. This degradation highlights the challenge of using a shared backbone, where competing gradients from different tasks interfere with each other during training. Despite its decoupled design, they cannot fully eliminate this conflict as it updates the shared backbone with contradictory objectives. In contrast, our method achieves identical performance whether trained jointly or separately for each task. This perfect decoupling stems from maximizing diffusion priors through specialized modules without the need to update the shared backbone.
\vspace{-1em}

\begin{table}[t!]
\centering
\resizebox{\columnwidth}{!}{
\scriptsize
\begin{tabular}{l|c|cc}
\toprule
\multirow{2}{*}{Method}  
   & Detection  
   & \multicolumn{2}{c}{Re-ID} \\
\cmidrule(lr){2-2} \cmidrule(lr){3-4}
   & AP & mAP & Top-1 \\
\midrule \midrule

DMRNet~\cite{DMRNet} (joint optimized) & 86.6\textbf{(-1.5)} & 93.2\textbf{(-0.4)}  & 94.2\textbf{(-0.8)}  \\
DMRNet~\cite{DMRNet} (detection only)  & 88.1 & - & -  \\
DMRNet~\cite{DMRNet} (re-ID only)      & -    & 93.6  & 95.0  \\
\midrule
DMRNet++~\cite{DMRNet++} (joint optimized)    & 88.3 \textbf{(-1.1)} & 94.4 \textbf{(-0.7)} &  95.5 \textbf{(-1.1)}    \\
DMRNet++~\cite{DMRNet++} (detection only)     & 89.4  &   -  &  -    \\
DMRNet++~\cite{DMRNet++} (re-ID only)         &   -  & 95.1  & 96.6   \\
\midrule
Ours (joint optimized)  & 90.9  & 97.8 & 98.4  \\
Ours (detection only)   & 90.9  & - & -  \\
Ours (re-ID only)       & -  & 97.8  & 98.4  \\
\bottomrule
\end{tabular}
}
\vspace{-0.8em}
\caption{Comparison of joint and individual task optimization on CUHK-SYSU~\cite{OIM} test set. Numbers in parentheses indicate performance drop from individual to joint optimization.}
\label{tab:joint optimization}
\vspace{-0.3em}
\end{table}

\begin{table}[t]
     \centering
     \scriptsize
     \renewcommand{\arraystretch}{1.0} 
     \begin{tabular}{ >{\centering\arraybackslash}p{0.5cm} | >{\centering\arraybackslash}p{2.5cm} | >{\centering\arraybackslash}p{1.3cm} | >{\centering\arraybackslash}p{1.3cm} } 
         \toprule
         Type & Combination & mAP & Top-1 \\
         \midrule \midrule
         
         (a) & Level 1 Only & 50.0  & 85.6  \\
         (b) & Level 2 Only & 58.0  & 89.4 \\
         (c) & Level 3 Only & 56.7  & 89.0  \\
         \midrule
         
         (d) & Level 0 + 2 & 56.1 & 86.9\\
         (e) & Level 1 + 2 & 60.3 & 89.7 \\
         (f) & Level 2 + 3 & \underline{61.3} & \textbf{91.2} \\
         (g) & \textbf{Level 1 + 2 + 3} & \textbf{62.0} & \underline{91.0} \\
         (h) & All levels combined & 60.5 & 90.8 \\
         \toprule
         
     \end{tabular}
     \vspace{-1.2em}
     \caption{Ablation study on feature map selection for re-ID on PRW~\cite{PRW} test set. (a)-(c) use the top 7 highest performing features from a single level, (d)-(f) utilize 4 from each level, (g) combines 1, 2, and 4, and (h) uses 2 from each level.}
     \label{tab:feature_selection}
     \vspace{-1.8em}
\end{table}

\paragraph{Feature selection.}
Selecting task-specific features for detection and re-ID requires careful consideration. While detection achieves strong performance with a single feature map, re-ID benefits from a thoughtful selection of multiple feature maps to capture richer discriminative information. Table.~\ref{tab:feature_selection} presents the impact of different feature selection strategies on re-ID performance. This experiment is based on the layer-wise analysis provided in the Suppl.~\textcolor{iccvblue}{J}. The results demonstrate that combining multi-scale features from multiple levels consistently outperforms single-level features. This may be attributed to the complementary information captured at different scales. Notably, level 0 features performed poorly due to their limited spatial resolution, which restricts their capacity to encode discriminative features. The optimal performance is achieved with (g), yielding 62.0\% mAP and 91.0\% Top-1 accuracy, which we adopt in our final configuration.

\begin{figure}[t!]
   \centering
   \includegraphics[width=0.99\linewidth]{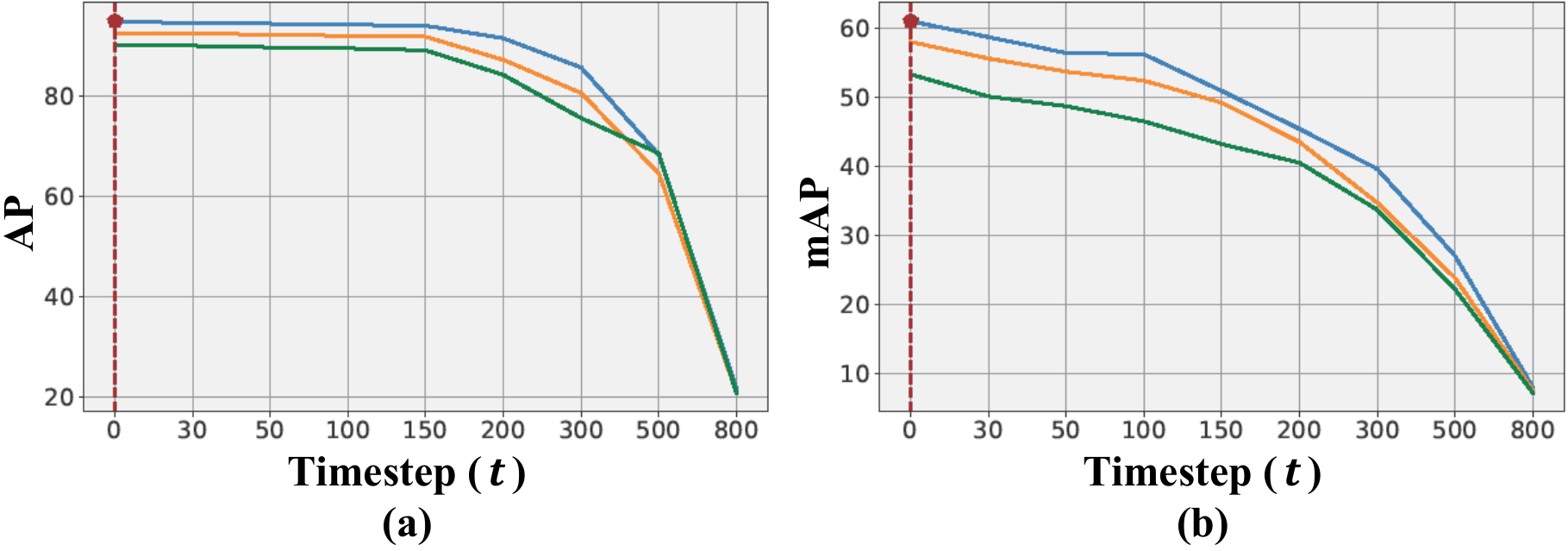}
   \vspace{-0.8em}
   \caption{Timestep analysis of diffusion features on PRW~\cite{PRW}. (a) Detection performance (AP) using three different layers from up-stage level 3 (blue, orange, green). (b) Re-ID performance (mAP) using features corresponding to (g), (b), and (c) from Table 3 (blue, orange, green). (Best viewed in color.)} 
   \label{fig:ablation-timestep}
   \vspace{-1.2em}
\end{figure}

\vspace{-1.0em}
\paragraph{Timesteps for detection and re-ID.}
As shown in Fig.~\ref{fig:ablation-timestep}, we observe distinct performance trends across different timesteps.  For detection (Fig.~\ref{fig:ablation-timestep}(a)), AP remains stable up to moderate timesteps but starts to decline beyond \textit{t}=200. This suggests that detection benefits from noise acting as a regularizer, improving robustness to varying scene conditions. In contrast, re-ID performance (mAP) deteriorates more rapidly, with a monotonic decline (Fig.~\ref{fig:ablation-timestep}(b)). Since re-ID relies on fine-grained details for identity discrimination, even small amounts of noise disrupt feature consistency, making it more sensitive to timestep variations. Notably, both tasks achieve optimal performance at \textit{t}=0, where the features preserve clear semantic structure without noise interference, supporting the analysis in Sec.~\ref{sec:Prior Knowledge of Diffusion Model}.
\vspace{-1.2em}

\paragraph{DGRPN.} We conduct an ablation study to evaluate the effectiveness of DGPRN, as shown in Table~\ref{tab:detection_ablation}. The baseline is Faster R-CNN~\cite{Faster-RCNN}, a widely used in prior methods~\cite{OIM, SeqNet, GFN, SEAS}, and we observe that incorporating DGPRN improves performance. This indicates that diffusion priors effectively guide region proposals by highlighting potential areas of interest. In the ablation study for \(\tau\), the results show that when \(\tau\) is too low, unnecessary regions are included, causing a drop in AP, while a high \(\tau\) removes valuable information, reducing recall. The optimal performance is achieved at \(\tau = 0.5\), balancing precision and recall.
\vspace{-1em}

\begin{figure}[t!]
   \centering
   \includegraphics[width=0.99\linewidth]{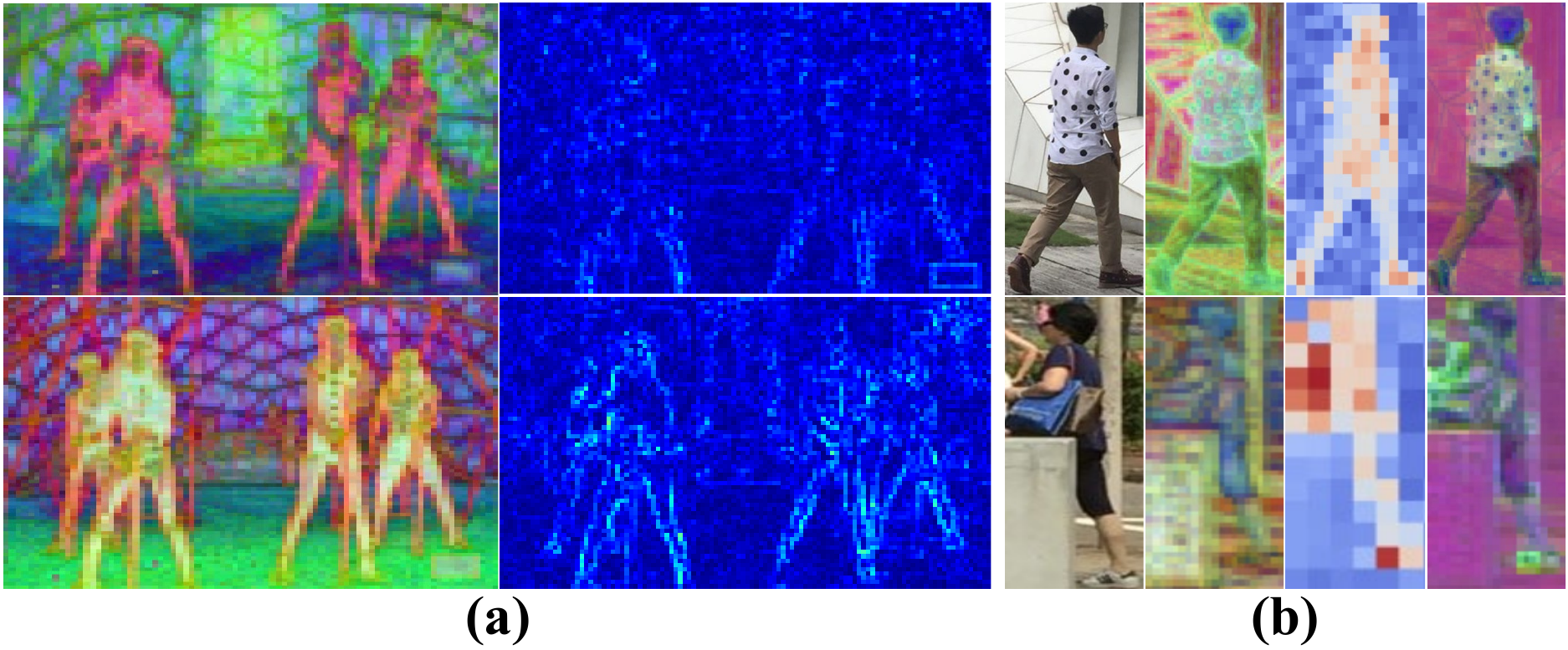}
   \vspace{-0.7em}
   \caption{Qualitative results of MSFRN and SFAN. (a) The first row shows features before MSFRN, while the second row presents the refined outputs, with the right column visualizing high-frequency components via DWT. (b) The first column displays original person crops, and the second column presents PCA~\cite{PCA} visualizations of features before being processed by SFAN. The third column shows the aggregated semantic maps \(\sum_{c} \hat{\mathbf{S}}_{c}\), and the last presents \(\mathbf{F}_{\text{sem}}\), the final semantic features.} 
   \label{fig:component}
   \vspace{-0.7em}
\end{figure}

\paragraph{MSFRN and SFAN.}
Table~\ref{tab:reid_ablation} presents the ablation study on PRW~\cite{PRW}, evaluating the impact of MSFRN and SFAN. The baseline employs DGRPN for detection and a part-based encoder for re-ID. Integrating MSFRN enhances mAP (+2.5\%) and Top-1 accuracy (+2.5\%) by refining high-frequency details, which are crucial for distinguishing similar individuals through identity-discriminative features. As shown in Fig.~\ref{fig:component}~(a), MSFRN effectively amplifies fine-grained patterns and high frequency information, as evident in the refined feature maps and corresponding high-frequency components. Adding SFAN further improves mAP (+0.5\%) and Top-1 accuracy (+0.3\%) by leveraging text-aligned semantic features, which enhance robustness to occlusions and pose variations through region-aware representations. Fig.~\ref{fig:component}~(b) illustrates how the semantic representation \(\hat{\mathbf{S}}_{c}\) highlights discriminative body parts, while the final refined feature \(\mathbf{F}_{\text{sem}}\) effectively suppresses background noise and occlusions, ensuring a stronger focus on the person. Further experiment on occlusion robustness is provided in the Suppl.~\textcolor{iccvblue}{F}. Combining MSFRN and SFAN in DiffPS yields the best performance (62.0\% mAP, 91.0\% Top-1), confirming that the two modules complement each other to improve re-ID performance.
\vspace{-0.4em}

\begin{table}[t]
    \centering
    \scriptsize
    \renewcommand{\arraystretch}{1.0}
    
    \begin{minipage}{0.43\columnwidth} 
    \centering
    \begin{tabular}{l|c|c}
        \toprule
        \textbf{Method} & \textbf{AP} & \textbf{Recall} \\
        \midrule
        Baseline & 94.2 & 97.5 \\
        + DGRPN & \textbf{94.8} & \textbf{98.1} \\
        \midrule
        \(\tau=0.2\) & 94.4 & 97.9 \\
        \(\tau=0.5\) & \textbf{94.8} & \textbf{98.1} \\
        \(\tau=0.8\) & 94.7 & 97.6 \\
        \bottomrule
    \end{tabular}
    \vspace{-0.8em}
    \caption{Ablation study of DGRPN on PRW~\cite{PRW}.}
    \label{tab:detection_ablation}
    \vspace{-1em}
\end{minipage}
\hfill 
\begin{minipage}{0.52\columnwidth} 
    \centering
    \begin{tabular}{c|c|c|c}
        \toprule
        \multicolumn{2}{c|}{\textbf{Components}} & \multicolumn{2}{c}{\textbf{Performance}} \\
        \cmidrule(lr){1-2} \cmidrule(lr){3-4} 
        \textbf{MSFRN} & \textbf{SFAN} & \textbf{mAP} & \textbf{Top-1}  \\
        \midrule
         &  & 59.1 & 88.1 \\
         \ding{51} &  & 61.6 & 90.6 \\
         & \ding{51} & 59.6 & 88.4  \\
        \ding{51} & \ding{51} & \textbf{62.0} & \textbf{91.0} \\
        \bottomrule
    \end{tabular}
    \vspace{-0.8em}
    \caption{Ablation study of MSFRN and SFAN on PRW~\cite{PRW}.}
    \label{tab:reid_ablation}
    \vspace{-1em}
\end{minipage}
\end{table}

\section{Conclusion}
In this paper, we propose DiffPS, a novel person search framework that fully leverages the prior knowledge of pre-trained diffusion models to address the fundamental challenges of backbone generalization and task conflict. Based on our analysis of diffusion priors, we introduce three specialized modules that maximize their effectiveness for person search. DiffPS completely resolves the optimization conflict between detection and re-ID while achieving state-of-the-art performance on CUHK-SYSU and PRW. 
\section*{Acknowledgements}
This work was supported by the National Research Foundation of Korea (NRF) grant funded by the Korea government (MSIT) (RS-2024-00355008). This research was also supported by the Culture, Sports and Tourism R\&D Program through the Korea Creative Content Agency grant funded by the Ministry of Culture, Sports and Tourism in 2024 (Project Name: Developing Professionals for R\&D in Contents Production Based on Generative AI and Cloud, Project Number: RS-2024-00352578, Contribution Rate: 25\%). Additionally, this work was supported by the National Research Foundation (NRF) grant (RS-2024-00421203), and by the Institute of Information \& Communications Technology Planning \& Evaluation (IITP) grant funded by the Korea government (MSIT) (RS-2021-II211343, Artificial Intelligence Graduate School Program (Seoul National University)).

{
    \small
    \bibliographystyle{ieeenat_fullname}
    \bibliography{main}
}

\clearpage
\setcounter{page}{1}
\maketitlesupplementary
\appendix

\section{UNet Architecture in Diffusion Models}
\label{supple:A}
The UNet~\cite{UNet} architecture in diffusion models follows a hierarchical structure, consisting of three primary stages: down-stage, mid-stage, and up-stage. Each of these stages is composed of multiple resolution levels, where feature activations at the same resolution are processed by a series of specialized modules, including ResNet~\cite{ResNet} blocks (Res blocks), Vision Transformer~\cite{ViT} blocks (ViT blocks), and up/down-samplers. These modules facilitate hierarchical feature extraction and enable efficient denoising by progressively reducing and restoring spatial resolution. The down-stage is responsible for reducing the spatial resolution of feature activations while increasing their channel depth. This stage comprises four resolution levels, with each level containing a sequence of Res blocks, ViT blocks, and down-samplers. The hierarchical nature of this stage allows the model to capture low-level details in the early layers and progressively extract more abstract and high-level features as the resolution decreases. At the lowest resolution, the mid-stage acts as a bottleneck layer that connects the down-stage and up-stage. It consists of stacked Res and ViT blocks, enabling feature refinement before upsampling begins. The up-stage mirrors the down-stage by progressively restoring spatial resolution through a sequence of Res blocks, ViT blocks, and up-samplers. Skip connections are established between corresponding levels in the down-stage and up-stage, allowing the network to propagate fine-grained details and prevent information loss.

\begin{table}[t!]
\footnotesize
\centering
\setlength{\tabcolsep}{2.5pt}
\begin{tabular}{l|l|cc|cc}
\toprule
\multirow{2}{*}{Detection Branch} & \multirow{2}{*}{Re-ID Branch} & \multicolumn{2}{c|}{Detection} & \multicolumn{2}{c}{Re-ID} \\
\cmidrule(lr){3-4} \cmidrule(lr){5-6}
 &  & Recall & AP & mAP & Top-1 \\
\midrule
Faster R-CNN~\cite{Faster-RCNN} & \multirow{2}{*}{Ours} & 97.6 & 94.5 & 61.8 & \underline{90.8} \\
RetinaNet~\cite{RetinaNet} &  & \underline{97.8} & \underline{94.6} & \underline{61.9} & \textbf{91.0} \\
\midrule
\multirow{4}{*}{Ours} & MGN~\cite{MGN} & \textbf{98.1} & \textbf{94.8} & 59.1 & 88.4 \\
 & PCB~\cite{PCB} & \textbf{98.1} & \textbf{94.8} & 60.4 & 90.1\\
 & NAE~\cite{NAE} & \textbf{98.1} & \textbf{94.8} & 60.0 & 89.1 \\
 & SEAS~\cite{SEAS} & \textbf{98.1} & \textbf{94.8} & 60.7 & 89.3 \\
\midrule
\multirow{1}{*}{Ours} & Ours & \textbf{98.1} & \textbf{94.8} & \textbf{62.0} & \textbf{91.0} \\
\bottomrule
\end{tabular}
\vspace{-0.5em}
\caption{Performance comparison of different detection and re-ID models on PRW~\cite{PRW} dataset. Numbers in bold indicate the best performance and underscored ones are the second best.}
\vspace{-1em}
\label{table:model_integration}
\end{table}

\section{Plug-and-Play Compatibility}
\label{supple:B}
In Table~\ref{table:model_integration}, we demonstrate competitiveness of our proposed modules with other state-of-the-arts detection modules (Faster R-CNN \cite{Faster-RCNN} and RetinaNet~\cite{RetinaNet}) and re-ID modules (MGN~\cite{MGN}, PCB~\cite{PCB}, NAE~\cite{NAE}, and SEAS~\cite{SEAS}). Our detection branch, guided by the proposed Diffusion-Guided Region Proposal Network (DGRPN), achieves the highest recall (98.1\%) and AP (94.8\%), outperforming Faster R-CNN (97.6\%, 94.5\%) and RetinaNet (97.8\%, 94.6\%). This highlights the effectiveness of DGRPN in enhancing person localization using cross-attention maps. Additionally, our re-ID branch consistently outperforms existing re-ID modules. While SEAS~\cite{SEAS} achieves a mAP of 60.7\% and Top-1 accuracy of 89.3\%, our method further improves the performance to 62.0\% mAP and 91.0\% Top-1 accuracy, demonstrating the benefits of our proposed modules in re-ID task.

\begin{table}[t!]
    \centering
    \scriptsize
    \setlength{\tabcolsep}{6pt}
    \renewcommand{\arraystretch}{1.1}
    \begin{tabular}{l|cc}
        \toprule
        \textbf{Text Prompts} & \textbf{mAP} & \textbf{Top-1} \\
        \midrule
        \texttt{"head"}, \texttt{"upper body"}, \texttt{"lower body"}, \texttt{"foot"} & 61.5 & 90.5 \\
        \texttt{"face"}, \texttt{"torso"}, \texttt{"legs"}, \texttt{"foot"} & \underline{61.7} & \underline{90.8} \\
        \texttt{"head"}, \texttt{"shirts"}, \texttt{"pants"}, \texttt{"shoes"} (Ours) & \textbf{62.0} & \textbf{91.0} \\
        \bottomrule
    \end{tabular}
    \vspace{-0.5em}
    \caption{Ablation study on different text prompts for SFAN on PRW~\cite{PRW}. Using clothing-related prompts (\texttt{"shirts"} and \texttt{"pants"}) provides more stable and distinctive cues, leading to the best re-ID performance. Numbers in bold indicate the best performance and underscored ones are the second best.}
    \label{tab:text_prompt_ablation}
    \vspace{-1.5em}
\end{table}

\section{Text prompt}
\label{supple:C}
To investigate the impact of different text prompts used in Semantic-adaptive feature aggregation network (SFAN), we conduct an ablation study by varying the predefined body-region text embeddings, as shown in Table~\ref{tab:text_prompt_ablation}. We compare three sets of prompts: (1) \texttt{"head"}, \texttt{"upper body"}, \texttt{"lower body"}, and \texttt{"foot"}, (2) \texttt{"face"}, \texttt{"torso"}, \texttt{"legs"}, and \texttt{"foot"}, and (3) \texttt{"head"}, \texttt{"shirts"}, \texttt{"pants"}, and \texttt{"shoes"}. The results indicate that the third configuration achieves the best performance, with the highest mAP and Top-1 accuracy. This improvement is attributed to the fact that \texttt{"shirts"} and \texttt{"pants"} explicitly correspond to clothing attributes, which are more stable and visually distinctive compared to \texttt{"upper body"} or \texttt{"torso"}, which may introduce ambiguity due to pose variations and occlusions. Similarly, \texttt{"shoes"} provide a clearer distinction than \texttt{"foot"}, as they often contain more discriminative patterns (\textit{e.g.}, color or style differences) that aid re-identification. In contrast, configurations (1) and (2) show degraded performance, likely due to their reliance on more generalized body descriptors that do not directly capture clothing details, leading to less discriminative spatial attention maps. These findings confirm that selecting text prompts that directly correspond to clothing-related features improves the effectiveness of SFAN in enhancing person representations.

\begin{table}[t!]
    \centering
    \footnotesize
    \setlength{\tabcolsep}{6pt} 
    \renewcommand{\arraystretch}{1.1} 
    \begin{tabular}{l|cc}
        \toprule
        \textbf{Agg Net.} & \multicolumn{2}{c}{\textbf{Re-ID}} \\
        \cmidrule(lr){2-3}
        & \textbf{mAP} & \textbf{Top-1} \\
        \midrule
        Hyperfeature~\cite{Hyperfeature} & \underline{60.9} & 90.2 \\
        CWA~\cite{diffpose} & 60.6 & \underline{90.8} \\
        Ours (MSFRN) & \textbf{62.0} & \textbf{91.0} \\
        \bottomrule
    \end{tabular}
    \vspace{-0.5em}
    \caption{Ablation study on various aggregation networks. Our proposed MSFRN achieves superior mAP and Top-1 accuracy. Numbers in bold indicate the best performance and underscored ones are the second best.}
    \label{tab:agg_net_comparison}
    \vspace{-0.5em}
\end{table}

\section{Feature aggregation network}
\label{supple:D}
We investigate the impact of different aggregation network architectures on person search performance, as shown in Table~\ref{tab:agg_net_comparison}. We compare our MSFRN against several existing networks, including Hyperfeature~\cite{Hyperfeature} (Res block-based) and CWA~\cite{diffpose}. Our proposed MSFRN, consisting of a multi-scale frequency refinement strategy, achieves superior performance with 62.0\% mAP and 91.0\% Top-1 accuracy, outperforming existing methods. This improvement stems from MSFRN's ability to effectively preserve high-frequency details while maintaining global feature coherence, enabling the extraction of more discriminative identity representations.

\begin{table}[t]
\footnotesize
\centering
\setlength{\tabcolsep}{7pt}
\begin{tabular}{l|cc|cc}
\toprule
\multirow{2}{*}{Backbone} & \multicolumn{2}{c|}{Detection} & \multicolumn{2}{c}{Re-ID} \\
\cmidrule(lr){2-3} \cmidrule(lr){4-5}
 & Recall & AP & mAP & Top-1 \\
\midrule
DINO~\cite{dinov2} ViT-B~\cite{ViT} & 75.2 & 70.4 & 33.5 & 66.1 \\
DINO~\cite{dinov2} ViT-L~\cite{ViT} & 81.3 & 76.5 & 36.1 & 72.8 \\
DINO~\cite{dinov2} ViT-G~\cite{ViT} & 84.5 & 79.8 & 41.5 & 76.8 \\
SD v1-5~\cite{StableDiffusion} & \underline{97.8} & \textbf{94.8} & \underline{61.3} & \underline{89.7} \\
SD v2-1~\cite{StableDiffusion} & \textbf{98.1} & \textbf{94.8} & \textbf{62.0} & \textbf{91.0} \\
\bottomrule
\end{tabular}
\vspace{-0.5em}
\caption{Comparison of different pre-trained frozen backbones in our framework. We compare Stable Diffusion~\cite{SD, StableDiffusion} (SD) v1-5 and v2-1 with DINO~\cite{DINO, dinov2} models of varying sizes (Base, Large, Giant) on the PRW~\cite{PRW} dataset. Numbers in bold indicate the best performance and underscored ones are the second best.}
\label{table:backbone_comparison}
\vspace{-1em}
\end{table}

\section{Pre-trained Backbone Selection}
\label{supple:E}
In Table~\ref{table:backbone_comparison}, we compare two different types of pre-trained foundation models as our backbone: DINO~\cite{dinov2}, trained via self-supervised learning, and Stable Diffusion (SD)~\cite{StableDiffusion}, trained through text-to-image generative modeling. We compare against DINO considering its strong performance in various visual recognition tasks. For fair comparison, we carefully configure DINO's feature extraction: the last layer token features are used for detection to leverage high-level semantic understanding, while features from the last seven layers are aggregated for re-ID. Our results show that SD significantly outperforms DINO variants across all metrics. While DINO learns to align representations between teacher and student networks, SD learns to reconstruct the complete visual hierarchy through the denoising process. The iterative denoising process of SD enables the model to learn both fine-grained appearance details and global structural information simultaneously, which naturally aligns with both requirements of person search. This comprehensive feature learning proves more effective than the instance-level discrimination of DINO, as evidenced by superior detection performance and re-ID accuracy.
\vspace{-0.5em}

\section{Key Challenges in Person Search}
\label{supple:F}
\begin{table}[t]
    \centering
    \footnotesize
    \begin{tabular}{l|cc}
        \toprule
        Method & \multicolumn{2}{c}{Re-ID} \\
        \cmidrule(lr){2-3}
        & mAP & Top-1 \\
        \midrule
        COAT\dag            & 86.5              & 85.6                \\
        SeqNeXt                     & 91.1              & 89.8                \\
        SeqNeXt+GFN                 & \underline{92.0}  & \underline{90.9}    \\
        SEAS\dag            & 89.6              & 87.7                \\
        Ours                        & \textbf{93.0}     & \textbf{91.9}       \\
        \bottomrule
    \end{tabular}
    \vspace{-0.5em}
    \caption{Occluded re-ID performance comparison across different methods on CUHK-SYSU~\cite{OIM}. Performance metrics using occluded person queries, demonstrating the effectiveness of our method under occlusion conditions. \dag 
 : Methods directly implemented or reproduced by us.}
    \label{tab:occlusion_comparison}
    \vspace{-1.7em}
\end{table}

\paragraph{Occluded person search.}
We show in Table~\ref{tab:occlusion_comparison} the robustness of our DiffPS to occlusion in person search. The evaluation protocol consists of 187 occluded person queries paired with a gallery of 50 images, where each query contains significant occlusion to simulate real-world scenarios. While occlusion poses a significant challenge in person search due to incomplete visual information, our framework achieves state-of-the-art performance (mAP = 93.0\%, Top-1 = 91.9\%) on the occluded person retrieval task. This superior performance under occlusion can be attributed to the generative nature of diffusion models, which learn to reconstruct complete visual information through the denoising process. This learned ability to recover missing or corrupted visual details enables our model to maintain robust person matching even when key body parts are occluded.
\vspace{-0.5em}

\begin{figure*}[t!]
\centering
\includegraphics[width=\linewidth]{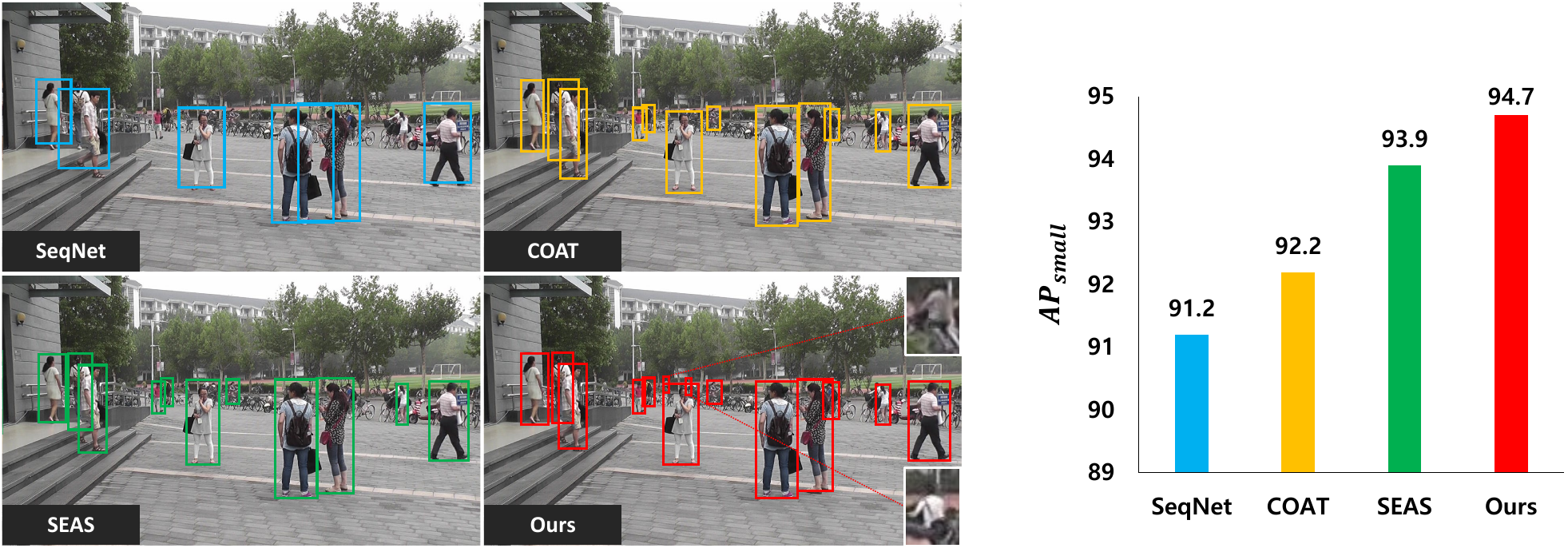}
\vspace{-1.8em}
\caption{Qualitative and quantitative comparison of small person detection performance. Left: Quantitative comparison of AP$_\text{small}$ scores on the PRW~\cite{PRW} test set across different methods, showing our model's superior performance in small person detection. Right: Visual comparison between SeqNet~\cite{SeqNet}, COAT~\cite{COAT}, SEAS~\cite{SEAS}, and our method on a challenging scene from the PRW test set containing multi-scale persons. Different colored boxes indicate detection results from each method.}
\label{fig:Detection_comparsion}
\vspace{-0.2em}
\end{figure*}

\begin{table*}[t!]
\centering
\renewcommand{\arraystretch}{0.9}
\scriptsize
\resizebox{0.7\textwidth}{!}{
\begin{tabular}{c|l|c|cc|cc}  
\toprule
 & \multirow{2}{*}{Method} & \multirow{2}{*}{Backbone} & \multicolumn{2}{c|}{PRW} & \multicolumn{2}{c}{CUHK-SYSU} \\
\cmidrule{4-5} \cmidrule{6-7}
 & & & Detection & Re-ID & Detection & Re-ID \\
\midrule
(a) & COAT & ResNet50 & 93.3 / 96.0 & 53.3 / 87.4 & 88.3 / 91.6 & 94.2 / 94.7 \\
(b) & COAT & SD v2-1 & 94.1 / 96.3 & 58.9 / 89.5 & 89.5 / 92.9 & 95.3 / 96.1 \\
\midrule
(c) & SEAS & ConvNeXt & 94.3 / 97.6 & 60.5 / 89.5 & 90.0 / 93.6 & 97.1 / 97.8 \\
(d) & SEAS & SD v2-1 & 94.5 / 97.5 & 60.8 / 90.1 & 90.3 / 93.9 & 97.3 / 97.7 \\
\midrule
(e) & Baseline (B) & SD v2-1 & 94.2 / 97.5 & 59.1 / 88.1 & 90.2 / 94.0 & 95.5 / 96.1 \\
(f) & B + D & SD v2-1 & \textbf{94.8} / \textbf{98.1} & 59.2 / 88.3 & \textbf{90.9} / \textbf{94.4} & 95.6 / 96.2\\
(g) & B + D + S & SD v2-1 & \textbf{94.8} / \textbf{98.1} & 59.6 / 88.5 & \textbf{90.9} / \textbf{94.4} & 96.4 / 96.8 \\
(h) & B + D + M & SD v2-1 & \textbf{94.8} / \textbf{98.1} & 61.6 / 90.6 & \textbf{90.9} / \textbf{94.4} & 97.0 / 97.8 \\
(i) & B + D + M + S & SD v2-1 & \textbf{94.8} / \textbf{98.1} & \textbf{62.0} / \textbf{91.0} & \textbf{90.9} / \textbf{94.4} & \textbf{97.8} / \textbf{98.4} \\
\bottomrule
\end{tabular}
}
\vspace{-0.5em}
\caption{D: DGRPN, M: MSFRN, S: SFAN. Detection is evaluated by AP / Recall, and Re-ID by mAP / Top-1.}
\vspace{-0.8em}
\label{tab_backbone}
\end{table*}

\paragraph{Small-scale person detection.}
Person search requires accurate person detection across various scales. While existing state-of-the-art methods~\cite{COAT, PSTR, GFN, SEAS} achieve strong performance on medium and large-scale persons, detecting small-scale persons remains a significant challenge. We demonstrate in Fig.~\ref{fig:Detection_comparsion} our model's superior capability in addressing this challenge. We define small-scale instances as those whose bounding box areas fall within the bottom 25\% of all bounding box areas in the dataset. As shown in the left of Fig.~\ref{fig:Detection_comparsion}, our framework achieves superior performance in small object detection (AP$_\text{small}$ = 94.7\%) compared to existing methods. This strong performance on small instances may stems from two key characteristics of diffusion models: 1) the iterative denoising process inherently requires the model to learn multi-scale feature representations, from fine details to global structures, making it particularly effective at capturing small object features; 2) diffusion models are trained on large-scale datasets with diverse scene compositions, enabling them to learn robust representations of objects at various scales and contexts. The qualitative comparison in the right of Fig.~\ref{fig:Detection_comparsion} clearly demonstrates this advantage. In a challenging scene with multiple small persons against a cluttered background, our proposed DiffPS demonstrates superior detection performance on small-scale persons compared to existing methods. This visual evidence indicates that the prior knowledge learned through generative modeling is particularly beneficial for challenging scenarios like small object detection, even without task-specific fine-tuning.

\section{Module Effectiveness}
\label{supple:G}
To rigorously validate the effectiveness of our proposed modules beyond the impact of the backbone itself, we conduct experiments using the same diffusion backbone across existing methods, as shown in Table~\ref{tab_backbone}. Specifically, rows (b), (d), and (i) demonstrate that even when applying SD v2-1 to existing frameworks, our method still achieves superior performance. This indicates that our performance gains are not simply due to the choice of a stronger backbone. Furthermore, row (e) presents a baseline that utilizes the SD v2-1 backbone without any of our proposed modules. Notably, this baseline performs worse than existing methods, highlighting that the backbone alone is insufficient to achieve state-of-the-art performance. From rows (e) to (i), we incorporate our proposed modules into the baseline, clearly showing that each module contributes meaningfully to performance improvement.

\section{Shape Bias}
\label{supple:H}
\begin{figure}[t!]
  \centering
  \includegraphics[width=0.99\linewidth]{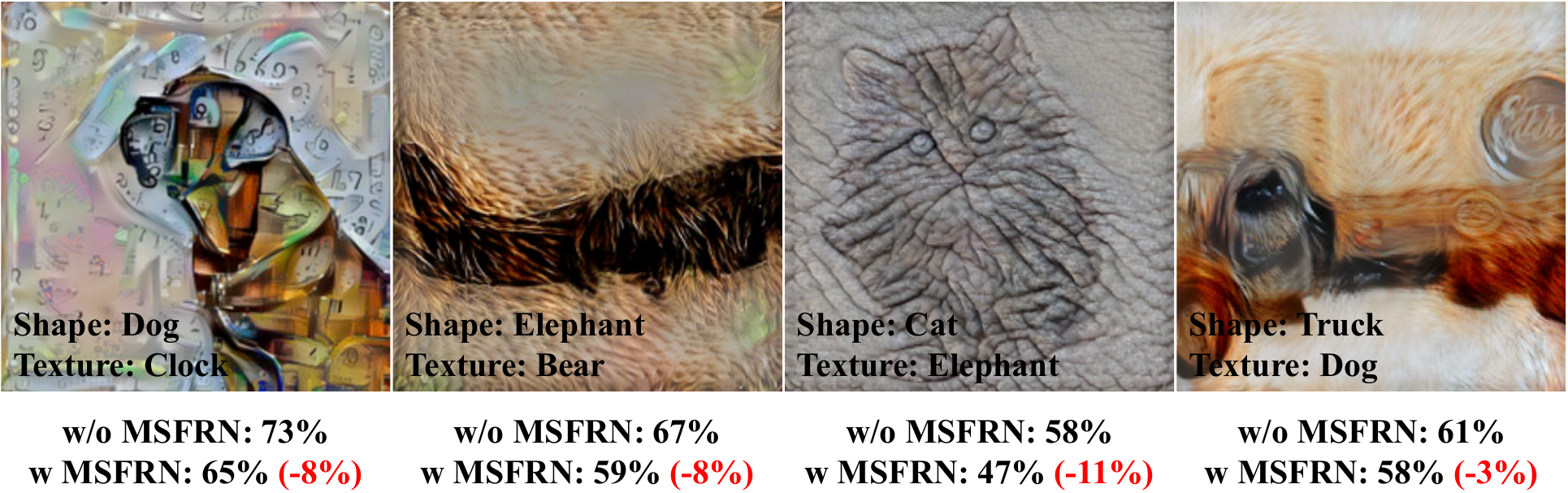}
  \vspace{-0.4em}
  \caption{Cue-conflict examples with shape/texture labels and model prediction probabilities with and without MSFRN.}
  \label{fig:shape bias}
\end{figure}

To directly validate that MSFRN mitigates shape bias, we conduct an experiment using the Cue-conflict~\cite{cueconflict} dataset, which is specifically designed to test whether a model relies more on shape or texture. As shown in Fig.~\ref{fig:shape bias}, this dataset contains images where the shape belongs to one class, but the texture is replaced with that of a different class. If the model predicts the label based on the shape, it means the model is biased toward shape information. For example, in the first image of Fig.~\ref{fig:shape bias}, where the shape corresponds to a dog and the texture to a clock, a shape-biased model would classify it as a dog. Table~\ref{tab:shape_bias} and Fig.~\ref{fig:shape bias} show the shape classification accuracy with and without MSFRN. Applying MSFRN reduces shape bias in both models, suggesting its effectiveness in reducing shape reliance and enhancing focus on fine-grained textures.

\section{Effect of $\delta$.}
\label{supple:I}
We investigate the effect of the hyperparameter $\delta$ in our Gaussian proposal mechanism within the Diffusion-Guided Region Proposal Network (DGRPN). $\delta$ controls the minimum spatial extent of the Gaussian standard deviation used to modulate attention-based proposals. As shown in our ablation study on the PRW dataset, both overly small and large $\delta$ values degrade performance: small values fail to suppress noisy or irrelevant regions, while large values over-smooth the localization map, reducing precision. The best performance is achieved at $\delta = 5$, which effectively balances precision and recall, leading to optimal detection performance.

\begin{table}[t!]
    \centering
    \footnotesize
    \begin{minipage}{0.48\columnwidth}
    \centering
    {\renewcommand{\arraystretch}{1} 
    \begin{tabular}{c|c}
        \toprule
        \textbf{Model} & \textbf{Shape ↓} \\
        \midrule
        ResNet50 & 28.18 \\
        + MSFRN  & \textbf{26.32} \\
        \midrule
        SD v2-1  & 63.52 \\
        + MSFRN  & \textbf{58.28} \\
        \bottomrule
    \end{tabular}
    \vspace{-0.5em}
    \caption{Shape bias mitigation.}
    \label{tab:shape_bias}
    }
    \vspace{-1.2em}
    \end{minipage}
    \hfill
    \begin{minipage}{0.45\columnwidth}
    \centering
    {\renewcommand{\arraystretch}{1.09}
    \begin{tabular}{c|c|c}
        \toprule
        \textbf{$\delta$} & \textbf{AP} & \textbf{Recall} \\
        \midrule
        1 & 94.3 & 97.6 \\
        3 & 94.7 & 97.9 \\
        5 & \textbf{94.8} & \textbf{98.1} \\
        7 & 94.6 & 97.7 \\
        \bottomrule
    \end{tabular}
    \vspace{-0.37em}
    \caption{Ablation study on $\delta$}
    \label{tab:delta_ablation}
    }
    \vspace{-1.2em}
    \end{minipage}
\end{table}


\renewcommand{\arraystretch}{0.9}
\begin{table}[t!]
\small
\centering
\setlength{\tabcolsep}{4pt}
\begin{tabular}{l|cc|cc}
\toprule
Layer & \multicolumn{2}{c|}{Detection} & \multicolumn{2}{c}{Re-ID} \\
\cmidrule(lr){2-3} \cmidrule(lr){4-5}
 & Recall & AP & mAP & Top-1 \\
\midrule
Down-stage Level0 Res0 & 95.4 & 91.1 & 42.1 & 83.2 \\
Down-stage Level0 ViT0 & 95.9 & 91.7 & 43.3 & 84.1 \\
Down-stage Level0 Res1 & 95.8 & 91.5 & 44.9 & 83.5 \\
Down-stage Level0 ViT1 & 95.6 & 81.3 & 44.3 & 84.0 \\
Down-stage Level0 Downsampler & 95.1 & 91.1 & 42.4 & 82.5 \\
\midrule
Down-stage Level1 Res0 & 96.0 & 82.4 & 43.6 & 83.1 \\
Down-stage Level1 ViT0 & 95.9 & 92.3 & 46.5 & 84.8 \\
Down-stage Level1 Res1 & \underline{96.2} & \underline{92.7} & \underline{47.3} & \underline{84.9} \\
Down-stage Level1 ViT1 & \textbf{96.3} & \textbf{92.9} & \textbf{48.7} & \textbf{85.6} \\
Down-stage Level1 Downsampler & 94.6 & 90.7 & 39.6 & 80.1 \\
\midrule
Down-stage Level2 Res0 & 95.0 & 91.3 & 42.7 & 81.2 \\
Down-stage Level2 ViT0 & 94.9 & 91.4 & 43.5 & 82.5 \\
Down-stage Level2 Res1 & 94.9 & 91.5 & 43.5 & 81.3 \\
Down-stage Level2 ViT1 & 95.3 & 92.1 & 41.9 & 81.6 \\
Down-stage Level2 Downsampler & 91.1 & 83.2 & 9.6 & 43.3 \\
\midrule
Down-stage Level3 Res0 & 91.1 & 82.8 & 8.4 & 40.4 \\
Down-stage Level3 Res1 & 90.0 & 81.7 & 6.9 & 36.4 \\
\midrule
Mid-stage Res0 & 90.2 & 81.5 & 6.4 & 33.7 \\
Mid-stage ViT0 & 91 & 81.8 & 6.4 & 34 \\
Mid-stage Res1 & 90.5 & 81.6 & 6.3 & 34.3 \\
\toprule
\end{tabular}
\vspace{-0.8em}
\caption{Performance metrics for different layers in the down-stage and mid-stage of UNet on the PRW~\cite{PRW} dataset. We evaluate different feature maps obtained from Vision Transformer~\cite{ViT} (ViT) and ResNet~\cite{ResNet} (Res) modules at each level. Each level contains multiple ViT and Res modules arranged sequentially, with the appended number (e.g., Res0, ViT0) indicating their order within that level. The downsampler represents feature maps from modules that reduce spatial resolution between adjacent levels. Numbers in bold indicate the best performance and underscored ones are the second best.}
\label{table:downandmidstage_layer}
\vspace{-1em}
\end{table}

\section{Analysis on Feature Map}
\label{supple:J}
\begin{figure*}[h!]
\centering
\includegraphics[width=\linewidth,height=0.8\textheight,keepaspectratio]{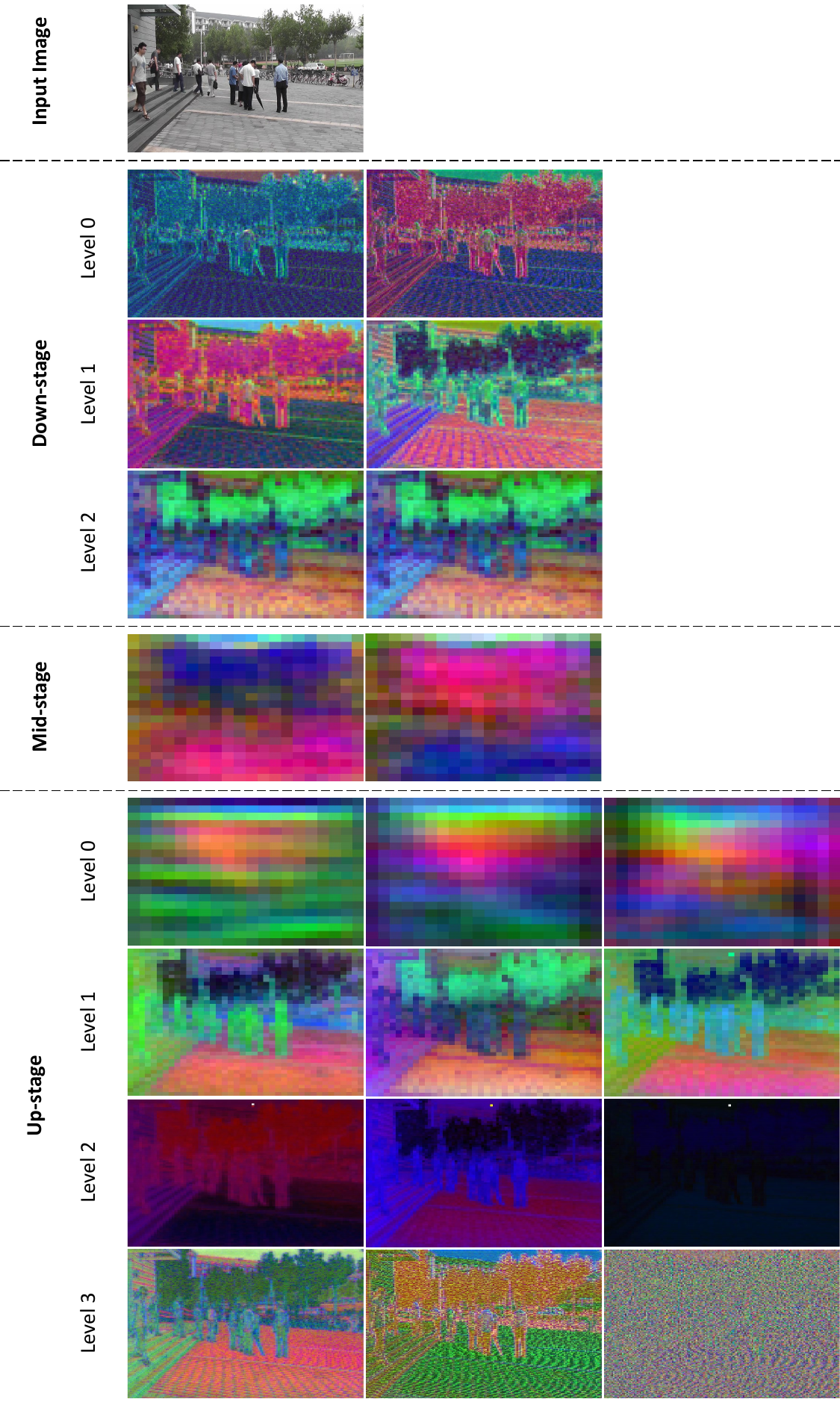}
\caption{Feature map visualization from Res~\cite{ResNet} modules across different stages and levels of UNet~\cite{UNet}. The visualizations are generated using PCA~\cite{PCA} on the output feature maps, with each row showing a different level and each column representing different res modules within that level. The input image is shown at the top for reference. Colors indicate the intensity and pattern of feature activations, demonstrating how feature representations evolve through different stages and levels of the network. (Best viewed in color.)}
\label{fig:Additional layer-resnet}
\end{figure*}

\paragraph{Layer-wise analysis}
We demonstrate feature characteristics of different layers and modules within the UNet~\cite{UNet} architecture through quantitative and qualitative analysis. As shown in Tables~\ref{table:downandmidstage_layer} and~\ref{table:upstage_layer}, the up-stage features consistently outperform their down-stage and mid-stage counterparts across all metrics. While down-stage features show moderate performance and mid-stage features demonstrate notably degraded performance, up-stage features exhibit remarkably superior performance, particularly in levels 2 and 3. The superior performance of up-stage features is further validated through qualitative analysis, which also reveals how different modules at the same level complement each other. Figure.~\ref{fig:Additional layer-resnet} shows that up-stage features from ResNet~\cite{ResNet} (Res) modules, especially at levels 2 and 3, maintain more distinctive patterns than their down-stage and mid-stage counterparts. This comprehensive analysis through both quantitative metrics and qualitative visualizations demonstrates that upper-level features in the up-stage possess strong discriminative power for person search.

\begin{figure*}[h!]
\centering
\includegraphics[width=\linewidth,height=0.8\textheight,keepaspectratio]{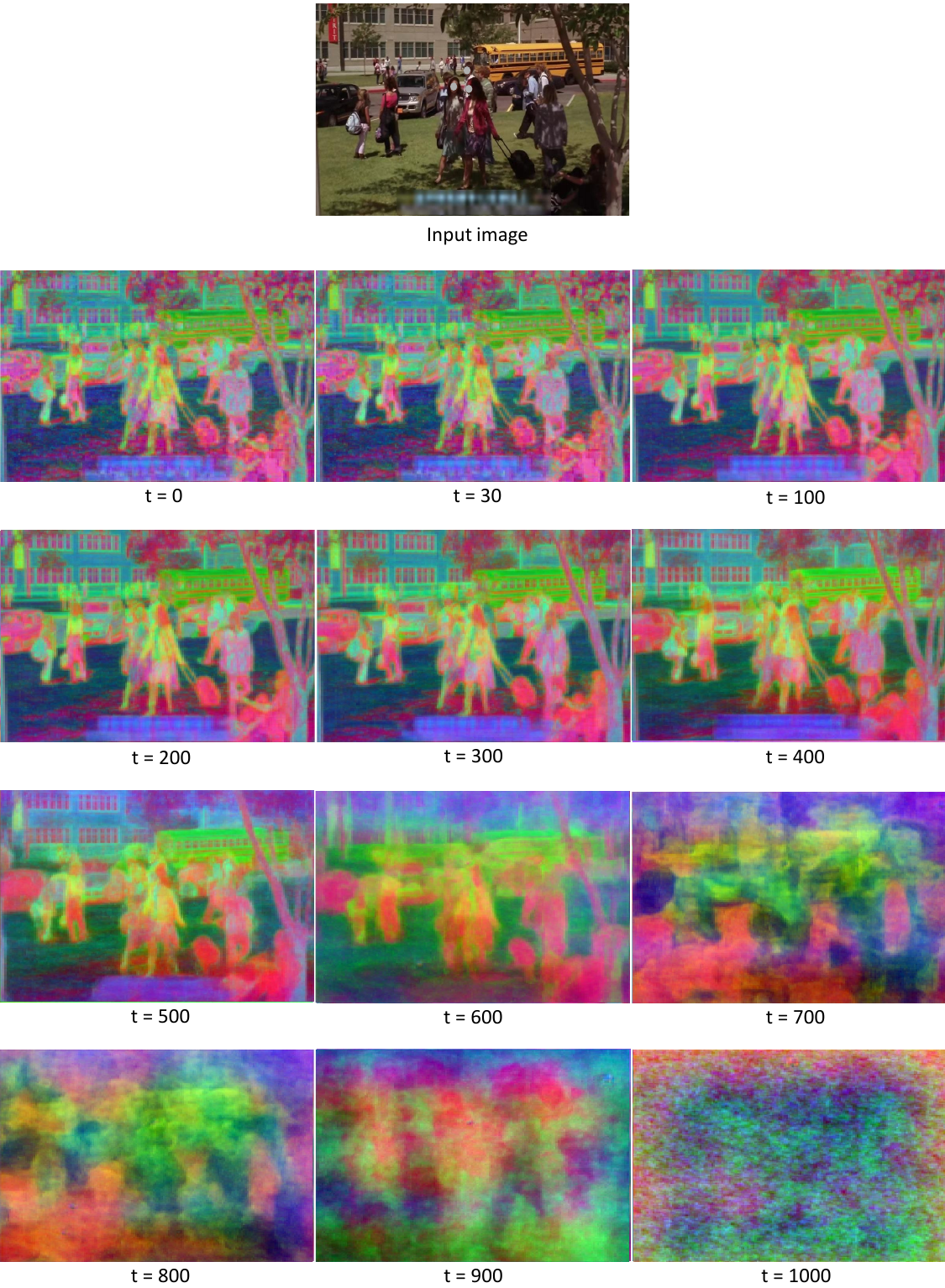}
\caption{Visualization of feature characteristics across different timesteps in the diffusion process. Visualization using PCA~\cite{PCA} of features extracted from UNet~\cite{UNet} up-stage level 3 ViT~\cite{ViT} module at varying timesteps. The input image is shown at the top for reference. (Best viewed in color.)}
\label{fig:Additional timestep}
\end{figure*}

\paragraph{Timestep-wise analysis}
We show in Fig.~\ref{fig:Additional timestep} how feature representations evolve across different timesteps \textit{(t)}. At \textit{t=0}, features maintain clear semantic structure with precise person silhouettes, leading to optimal re-ID and detection performance. Features gradually degrade through intermediate timesteps (\textit{t}=100-400), with person silhouettes becoming increasingly abstract. Later timesteps (\textit{t}=500-1000) show severe degradation, with features becoming dominated by noise and losing meaningful patterns. Figure~\ref{fig:Additional timestep} shows this progression in detail. This analysis reveals that early timesteps (\textit{t}=0-30) provide the most effective features for re-ID and detection tasks, informing our optimal timestep selection.


\renewcommand{\arraystretch}{0.85}
\begin{table}[h!]
\small
\centering
\setlength{\tabcolsep}{4pt}
\begin{tabular}{l|cc|cc}
\toprule
Layer & \multicolumn{2}{c|}{Detection} & \multicolumn{2}{c}{Re-ID} \\
\cmidrule(lr){2-3} \cmidrule(lr){4-5}
 & Recall & AP & mAP & Top-1 \\
\midrule
Up-stage Level0 Res0 & 88.3 & 76.5 & 1.5 & 11.3 \\
Up-stage Level0 Res1 & 89.5 & 78.3 & 1.6 & 12.2 \\
Up-stage Level0 Res2 & 89.2 & 79.2 & 1.8 & 14.1 \\
Up-stage Level0 Upsampler & 88.7 & 79.4 & 1.1 & 8.9 \\
\midrule
Up-stage Level1 Res0 & 96.0 & 92.7 & 41.6 & 80.9 \\
Up-stage Level1 ViT0 query & 95.3 & 91.6 & 40.2 & 80.0 \\
Up-stage Level1 ViT0 key & 95.2 & 91.7 & 40.6 & 79.6 \\
Up-stage Level1 ViT0 value & 95.4 & 51.8 & 39.6 & 79.0 \\
Up-stage Level1 ViT0 & 95.3 & 91.5 & 37.7 & 79.0 \\
Up-stage Level1 Res1 & 96.1 & 92.7 & 44.9 & 82.6 \\
Up-stage Level1 ViT1 query & 95.8 & 92.6 & 42.3 & 81.2 \\
Up-stage Level1 ViT1 key & 95.9 & 92.4 & 41.5 & 80.5 \\
Up-stage Level1 ViT1 value & 95.7 & 92.3 & 42.7 & 81.7 \\
Up-stage Level1 ViT1 & 95.9 & 92.6 & 40.7 & 80.6 \\
Up-stage Level1 Res2 & 96.0 & 92.8 & 46.3 & 83.0 \\
Up-stage Level1 ViT2 query & 95.8 & 92.5 & 46.4 & 83.6 \\
Up-stage Level1 ViT2 key & 95.5 & 92.2 & 46.0 & 82.8 \\
Up-stage Level1 ViT2 value & 95.2 & 91.6 & 45.8 & 82.8 \\
Up-stage Level1 ViT2 & 95.5 & 92.2 & 45.1 & 82.8 \\
Up-stage Level1 Upsampler & 96.8 & 93.7 & 39.4 & 80.5 \\
\midrule
Up-stage Level2 Res0 & 97.4 & 94.3 & 50.4 & 86.7 \\
Up-stage Level2 ViT0 query & 97.7 & \underline{94.7} & 50.0 & 85.2 \\
Up-stage Level2 ViT0 key & 97.5 & 94.5 & 48.7 & 84.3 \\
Up-stage Level2 ViT0 value & 97.2 & 94.3 & 48.5 & 85.2 \\
Up-stage Level2 ViT0 & 97.2 & 94.3 & 47.3 & 84.1 \\
Up-stage Level2 Res1 & 97.6 & 94.6 & 53.7 & 86.4 \\
Up-stage Level2 ViT1 query & 97.6 & 94.5 & 53.6 & 85.9 \\
Up-stage Level2 ViT1 key & 97.6 & 94.5 & 52.3 & 85.7 \\
Up-stage Level2 ViT1 value & 97.5 & 94.1 & \underline{53.8} & 87.1 \\
Up-stage Level2 ViT1 & 97.3 & 94.2 & 52.1 & 86.5 \\
Up-stage Level2 Res2 & 97.4 & 94.4 & 53.5 & \textbf{87.9} \\
Up-stage Level2 ViT2 query & 97.3 & 94.3 & \textbf{54.4} & 87.3 \\
Up-stage Level2 ViT2 key & 97.8 & 94.5 & 53.5 & 86.5 \\
Up-stage Level2 ViT2 value & 96.9 & 93.8 & 52.7 & 86.3 \\
Up-stage Level2 ViT2 & 97.2 & 94.2 & 51.9 & 86.4 \\
Up-stage Level2 Upsampler & 97.8 & \underline{94.7} & 51.9 & 86.5 \\
\midrule
Up-stage Level3 Res0 & 97.4 & 94.1 & 52.2 & 86.4 \\
Up-stage Level3 ViT0 query & \underline{98.0} & \underline{94.7} & 52.9 & 87.3 \\
Up-stage Level3 ViT0 key & \textbf{98.1} & \textbf{94.8} & 53.1 & \underline{87.7} \\
Up-stage Level3 ViT0 value & 97.4 & 94.0 & 53.1 & 87.0 \\
Up-stage Level3 ViT0 & 97.1 & 93.6 & 47.1 & 84.8 \\
Up-stage Level3 Res1 & 97.7 & 94.3 & 52.0 & 86.2 \\
Up-stage Level3 ViT1 query & 97.5 & 94.1 & 52.8 & 86.6 \\
Up-stage Level3 ViT1 key & 97.5 & 94.2 & 53.2 & 86.8 \\
Up-stage Level3 ViT1 value & 97.5 & 94.0 & 52.7 & 86.3 \\
Up-stage Level3 ViT1 & 97.4 & 93.8 & 48.1 & 85.0 \\
Up-stage Level3 Res2 & 97.1 & 93.6 & 48.2 & 85.2 \\
Up-stage Level3 ViT2 query & 97.1 & 93.9 & 51.1 & 86.7 \\
Up-stage Level3 ViT2 key & 97.4 & 94.1 & 51.5 & 86.0 \\
Up-stage Level3 ViT2 value & 96.5 & 92.6 & 47.1 & 84.3 \\
Up-stage Level3 ViT2 & 97.0 & 93.3 & 47.0 & 85.0 \\
\toprule
\end{tabular}
\vspace{-0.8em}
\caption{Performance metrics for different layers in the up-stage of UNet on the PRW~\cite{PRW} dataset. We evaluate feature maps from Vision Transformer~\cite{ViT} (ViT) and ResNet~\cite{ResNet} (Res) modules at each level. Each level contains multiple ViT and Res modules in sequence, with the appended number (\textit{e.g.}, Res0, ViT0) indicating their order. For ViT modules, we analyze three attention-based feature maps (query, key, and value) after their linear projections. The upsampler represents feature maps from modules that increase spatial resolution between adjacent levels. Numbers in bold indicate the best performance and underscored ones are the second best.}
\label{table:upstage_layer}
\end{table}

\section{Limitation}
In this work, we harness diffusion priors to person search and demonstrate their effectiveness. Our DiffPS leverages a pre-trained diffusion model as a large-scale foundation model, which could raise concerns about computational overhead. However, by adopting a frozen backbone, we maintain fewer learnable parameters compared to recent state-of-the-art models. Future research on efficient diffusion models could further address computational considerations while retaining our method's advantages.

\clearpage

\end{document}